\documentclass[11pt]{report}
\usepackage{tabularx}
\usepackage{graphicx} 
\usepackage{subcaption} 
\usepackage{lenovo}		
\usepackage[utf8]{inputenc}	
\usepackage[T1]{fontenc}
\usepackage{textcomp}		
\usepackage{amsmath}		
\usepackage{fancybox}
\usepackage{anyfontsize}	
\usepackage{lipsum}
\usepackage{float}
\usepackage[normalem]{ulem}
\usepackage{cleveref}
\usepackage{xcolor}
\usepackage{pdfpages}
\usepackage{geometry}
\usepackage[printonlyused]{acronym}
\hypersetup{
	pdftitle={Report Version: 0.001},
	pdfauthor={[AI Lab, Lenovo Research]},
	pdfkeywords={Technical Report, Lenovo (Beijing) Ltd., 2024.},
}

\graphicspath{ {graphics/} }

\def\hyperride{HYPERRIDE}
\def\letsfm{Le-TSFM} 
\def\leforecast{LeForecast} 
\def\repotitle{LeForecast: Enterprise Hybrid Forecast by Time Series Intelligence} 

\begin{document}

	
\urlstyle{same}


\ProjectFullTitle{LeForecast}
\ProjectAcronym{LeForecast}
\ProjectRefNo{xxxx}
\delivNumber{0.001}
\delivName{\repotitle{}}
\delivShortTile{LeForecast: Enterprise Hybrid Forecast Platform}
\delivResponsible{AI Lab, Lenovo Research, Lenovo (Beijing) Ltd.} 
\delivVersion{2024.10}
\delivFrom{01/04/2023}
\delivTo{31/12/2024}

\delivNumber{2}
\ActualDate{31/10/2024}
\delivDissLevel{CO}
\delivType{r}
\delivAuthor{LeForecast Team (leforecast@lenovo.com)}
\delivFPAuthor{Please refer to Page X for a list of authors}
\delivStatus{d}
\delivKeywords{[Technical Report]}
\delivStatus{Draft}
\delivExecSummary{This is a summary of the deliverable; a paragraph or
so to go on the cover page}


\makecover
\begin{abstract}

Demand is spiking in industrial fields for multidisciplinary forecasting, where a broad spectrum of sectors needs planning and forecasts to streamline intelligent business management, such as demand forecasting, product planning, inventory optimization, etc. Specifically, these tasks expecting intelligent approaches to learn from sequentially collected historical data and then foresee most possible trend, i.e. time series forecasting. Challenge of it lies in interpreting complex business contexts and the efficiency and generalisation of modelling. With aspirations of pre-trained foundational models for such purpose, given their remarkable success of large foundation model across legions of tasks, we disseminate \leforecast{}, an enterprise intelligence platform tailored for time series tasks. It integrates advanced interpretations of time series data and multi-source information, and a three-pillar modelling engine combining a large foundation model (Le-TSFM), multimodal model and hybrid model to derive insights, predict or infer futures, and then drive optimisation across multiple sectors in enterprise operations. The framework is composed by a model pool, model profiling module, and two different fusion approaches regarding original model architectures. Experimental results verify the efficiency of our trail fusion concepts: router-based fusion network and coordination of large and small models, resulting in high costs for redundant development and maintenance of models. This work reviews deployment of \leforecast{} and its performance in three industrial use cases. Our comprehensive experiments indicate that \leforecast{} is a profound and practical platform that enhances efficient and competitive performance. And we do hope that this work can enlighten the research and grounding of time series techniques in accelerating enterprise development. 
\end{abstract}

\clearpage
\fancypagestyle{plain}{}
\settableofcontents
\tableofcontents









\clearpage
\section{Introduction}
\label{sec:intro}

Rapid advancements in intelligent transformation have enhanced the efficiency of a multitude of business operations in past years: from procurement arrangement to product planning, from understanding patterns of energy consumption to economic cycles in finance, from capacity management in logistics and manufactories to controlling traffic flows. 
One of pivotal techniques for this transformation is efficient time series forecasting driven by machine learning (ML) on a large scale of data, to anticipate, strategise, and mitigate rapid changes and uncertainties in real-world for rational decision making. 
Time series forecasting refers to predicting future based on historical data that are collected or recorded at a continuous and regular pace of time. 
However, due to the diversity and complexity in business scenarios, the adoption of ML-based time series forecasting approaches in business mainly faces three critical challenges: problem modelling, knowledge injection, and maintenance costs. 
This work presents our efforts in building an enterprise platform that systemically integrates intelligent technologies for various time series forecasting tasks, consisting advanced interpretations of time series data and multi-source information, and a three-pillar modelling engine combining foundation model, multimodal model and hybrid model  to derive insights, predict or infer futures, and then drive optimisation  across multiple sectors in enterprise operations, namely time series intelligence.

\subsection{Time Series Forecasting}
In general, time series refers to a sequence of data points collected at consistent and successive intervals over a period of time. 
It usually showcases trends, seasonalities, cycles and variations, which feature temporal dependency, spatio dependency and semantic diversity \cite{Ye2024ASO}. Forecasting is one of essential tasks on top of time series, along with anomaly detection, classification, segmentation and so forth, which are the fundamental work in vast enterprise operations, such as transportation, supply chain, healthcare, finance, retail, etc. 
Example sectors include forecasts of product demands for efficient retail operation, traffic forecasts to utilise transportation resources, infectious disease forecasting for pandemic control, to name a few. 
These areas share the enthusiasm for accurate time series forecasting to innovate operations.

Research of time series forecasting has a long-standing history closely joined with real-world tasks. 
The first methodological analysis task on record goes back to weather forecasts in the 19th century \cite{fitzroy1863weather}. 
Since then, statistical approaches dominate the field of time series analysis, accelerating the development of various classic models, such as Exponential Smoothing and ARIMA, that are still widely used in numerous domains. 
In the past decades, the boom in data scale and computing power prompts advents of innovative approaches, especially ML and deep learning (DL) based models, including RNN, LSTM, Gradient Boosting, etc. 

These models are at an advantage to perform accurate forecasts over statistical approaches. 
However, one contrasting stand is that their performance suffers from scarce well-labelled data for real-world tasks, with the addition of extensive effort into model training, fine-tuning and customisation for every specific task. 
And scaling the forecasting capability outside the scope of their training data is impractical \cite{Darlow2024Dam}. 
In practice, marked distinctions across business sectors and significant variations in characteristics of time series data, which make it impossible to transfer forecasting models among different scenarios, but require customisation in accordance with vast variances in business preferences. 
This results in substantial costs associated with development and maintenance of both data and model.

\subsection{Towards Foundation Model}
Aspiration to address above limitations and challenges via foundational generative models have been boosted in the field of time series forecasting under the overwhelming impression of language foundation models' soaring potentials. Such models leverage the advanced techniques for transfer learning and self-supervision, i.e. pre-training models on source domain with a wide spectrum of data and then fine-tuning the models for tasks in similar domains. Practitioners in industry and academia underlie a presumption of a pre-trained foundation model tailored for time series that forecasts in a more general, accessible, accurate, time-consuming and practical manner. They have escalates restlessly development of foundation models for time series tasks, namely time series foundation models, implying that such models are pre-trained on time series data at great depth and breadth, presumably developing a deep understanding of underlying patterns in the data, and then can forecast accurately for unseen time series tasks across domains, with minimal effort in additional training or fine-tuning. 

The advantages of time series foundation model can be seen from recent works building upon transformer-based methodologies \cite{garza2023timegpt}, diffusion-based methodologies \cite{ansari2024chronos} and others \cite{rasul2023lag}. 
Specifically, they demonstrate comprehensive and adaptable forecasting capability for different use cases across domains of energy, economics, health, retail, etc. 
Furthermore, they can be used effectively for various tasks, such as anomaly detection and classification, and for different prediction requirements, such as long-term and short-term forecasting. 

Thereby, we summarise the following \textit {skill sets of forecasting models}, which could potentially offer advantages and specialisations tailored to cross-domain tasks at full-scale.
\begin{itemize}
\setlength\itemsep{-0.6em}
    \item Trend Sensitivity: Ability to detect and model trends.
    \item Seasonality Sensitivity: Ability to detect and model periodic patterns.
    \item State Transition Speed: Ability to adapt to sudden changes in patterns.
    \item High-entropy Forecasting: Ability to handle highly stochastic or noisy data.
    \item Short-term and Long-term Forecasting: Ability to forecast for both short and long horizons.
    \item Long-Term Memory: Ability to retain and utilize long-term dependencies in data.
    \item Intermittent Forecasting: Ability to predict sparse or irregular patterns effectively.
\end{itemize} 
While pre-training our time series foundation model, we continuously monitor and evaluate its capability with respect to the above suite. This enables targeted adjustments of our foundation model, ultimately contributing to its balanced performance across various dimension.

\subsection{Challenges of Enterprise Grounding}
Although time series foundation models are an answer to the limitations and challenges that statistic models and classic machine learning models can not overcome, grounding these models to optimise enterprise operations is not practical yet. 
First, enterprise tasks are usually associated with a broad landscape of information, including multiple information sources and modality. 
For instance, inventory management is one of typical operations in supply chain, which deeply relies on forecasting the amount of loading in and out with respect to status of sales market and manufactory planning. 
In such a scenario, an informed forecast shall learn historical data, align with business experiences, and sense external market status as well. 
However, massive amount of business experiences and expertise are unstructured or even undocumented, which need arduous work to impart and formalise. 
Additionally, despite the pivotal role of acquiring information from external sources in business, existing models are still inadequate for collecting and consuming real-time updates to deliver informed predictions. 

On the other hand, marked distinctions across business sectors weaken the generalization capability of time series foundation models. 
One of the reasons is that the patterns of business operations underlie different time series patterns. 
In our practice, logistics and retails apparently prefer divergent interpretations of time series. 
Logistics pays attention to the impact of extreme weathers and holidays on delivery times, whereas retails would give weight to stock indexes, competitors and live news. 
Thus, time series from those scenarios reflects the influence of different factors, thereby, persisting dissimilar patterns. 
This prevents time series foundation models from producing efficient forecasts across domains.
Contrastively, smaller models customised for specific business scenarios tend to outperform foundation models in this regard.

We're looking for technologies to generate forecasts. These forecasts should blend data, knowledge, and multiple models while collaborating with humans. 
This is to deliver business value, ensuring that the forecasts function reliably, nimbly, and trustworthily across diverse scenarios.

\subsection{Time Series Intelligence}

With aspirations to address the above challenges based on foundation models, we disseminate \leforecast{}, an enterprise hybrid forecast platform featuring time series intelligence, consisting of robust foundational forecasting and proactive contextual knowledge integrating capabilities across different scenarios. 
\leforecast{} delivers our concept, "Intelligence as Service", into a technological ecosystem to empower enterprises, optimise management processes and allocate resources.

Systematically, \leforecast{} first categorises the challenges into two classes: information and modelling, i.e. management of scenario-based data and knowledge as well as forecasting efficiency associated with model deployment and maintenance across multiple business scenarios, and then proposes initiatives and solutions to meet the challenges and establish a generalised time series forecasting capability for enterprises. 
Specifically, \leforecast{} constructs data governance and knowledge sensing at information landscape to profoundly integrate, augment and fuse valuable information from multiple sources. Building upon the information landscape, \leforecast{} assembles a three-pillar modelling layer with a time series foundation model, namely \letsfm{} (A Time Series Foundation Model), multimodal model and unified model fusion to deliver hybrid forecasts across domains. 

\subsubsection{Blending Information Landscape}
The very first difficulty that we encounter in dealing with real-world tasks usually lies in processing information, due to diverse formats and various qualities of complex sources. 
Before designating solutions, it is of fundamental importance that such kinds of information from multiple sources should be well managed into valuable data and knowledge regarding the nature of models and forecasting tasks. To this end, we separate the information landscape into time series data and contextual knowledge, thereby, propose the following two modules.
\begin{itemize}
    \item \textbf{Data Governance}: Regarding time series data from multiple information sources, we design a data governance module to manage the quality and integrity of data to support pre-training, monitoring, and validating time series forecasting capabilities \cite{Zha2023DatacentricAI}. To keep in line with existing work in the field of time series foundation models, data governance starts by collecting public time series data as long as we can access, then, extends breadth and depth of them via data augmentation and data optimisation to enhance the generalisation capability of \letsfm{} in pre-training. 
    
    \item \textbf{Information Mining}: \leforecast{} can forecast in the immediacy of live knowledge via mining external information. Regarding the source and format of information, it categorises 3 types of information sources, live news, structured data and business reports, and then proactively acquires insights from these sources through information sensors designated for each type of sources, respectively. In addition, it uses a separate sensor as an analyst to fuse those multi-source insights for consistency, accuracy and relevancy. To be precise, information mining leverages generative language model to collect, analyse and then summarise live status and trends from external information, as extra inputs to enrich the contextual knowledge for forecasting models to make more accurate and realistic predictions. 
\end{itemize}

\subsubsection{Three-Pillar Modelling}
Our initiatives of time series intelligence at enterprise level are to boost time series forecasting with the remarkable ability of foundation models, extend model intelligence through  multimodality learning and transfer general forecasting capabilities across domains without the losing effectiveness of smaller models tailored for specific domain tasks. 
To fulfil the objective hereof, the intelligence of \leforecast{} is further structured into a three-pillar modelling layer: time series foundation model, multimodal model and model fusion. 

\begin{itemize}
    \item \textbf{TSFM for Generalised Forecasting Capability}: The basic idea to enhance a transformer model's capability of generalised forecasting is to pre-train the model on a diverse dataset to learn insights across broad domains, thereby to improve the forecasting performance on unprecedented scenarios. \letsfm{} is a \textit{multi-expert decoder-only Transformer model}, pre-trained on extensive real-world data from 10 domains as well as synthetic data provided by the data governance module. 
    Such training datasets boost its robust generalization capability and adaptivity across business scenarios in two ways: zero-shot forecasting and fine-tuning, where zero-shot forecasting refers to solving new forecasting tasks without any modification of model parameters, while fine-tuning, conversely, implies training the parameters for identical tasks starting from the pre-trained ones. 

    \item \textbf{Multimodal Model for Multi-modal Learning Capability}: To mimic the cognitive ability of human in learning from multimodal knowledge, which, in the field of time series, implies that, \leforecast{} shall inherit advantages of language models in understanding linguistic contexts of time series while learning temporal patterns, and fuse its predictions with respect to real-time information. Three cross-modal alignment techniques contribute to the multi - model understanding ability: 
    \begin{enumerate}
        \item Historical Text-Time Series Alignment Loss: aligns descriptive historical text and corresponding time series data to strengthen the model's ability to learn inter-variable relationships.
        \item Historical-oriented Modality Interaction Module: encodes multimodal text-time series representations, ensuring effective alignment of distributions between historical text and time series data. 
        \item Future-oriented Modality Interaction Module: employs cross-attention to incorporate predictive textual insights, ensuring textual insights-following forecasting for more reasonable forecasts.
    \end{enumerate}
    
    \item \textbf{Module Fusion for Hybrid Forecasting Capability}: Instead of 'learning from data', model fusion enables the promising paradigm of 'learning from model'. We propose a framework to fuse various type of time series models, such as foundation models, small-scale deep learning models, statistic models, etc, into a single, cohesive model that leverages complementary potentials of individual time series models and obtains superior forecasting performance. Specifically, the framework is composed by a model pool, model profiling module, and two different fusion approaches regarding original model architectures. Experimental results verify the efficiency of our trail fusion concepts: router-based fusion network and coordination of large and small models. 
\end{itemize}

\subsection{Reliable Enterprise Adoption}
We ground \leforecast{} in real business scenarios and select the following three scenarios to showcase its commercial values and potentials:
\begin{itemize}
    \item \textbf{PC Demand Forecasting}: Accurate forecasts of demand on personal computers (PCs) are essential for Global E-Commerce to manage inventory, manufactories and supplements. Forecasts based on human experiences encapsulate market expectations, business principles and company strategies, but they lack effectiveness, accuracy and generality. \leforecast{} transfers such expertises via multimodal analysis into automatic and informed demand forecasting for Global E-Commerce, since then, they haven seen an increase of 5-7\% in profit.

    \item \textbf{Global Logistics Volume Forecasting}: They supports logistics operations to book transportation resources for global orders 7 days in advance. Inaccurate forecasts result in either higher costs or extended time of delivery i.e. bad customer experiences. \leforecast{} streamlines shipment booking by uplifting 20\% accuracy in shipment volume prediction, incorporating with shipment histories, manufacture plans and shipment schedules.
    
    \item \textbf{Factory Carbon Emissions Forecasting}: Companies have been prioritising ESG in their strategies, and cutting carbon emission across manufactories. 
    One of lighthouse factories utilises \leforecast{} to predict energy consumption and carbon emission for its operation and production. By fusing \leforecast{} and mathematical models, the factory delivers the very first systematic carbon emission prediction at above 80\% accuracy which yields a zero-carbon certification and recognition across the industry.
\end{itemize}

In the rest of this work: 
Section \ref{sec:data_goven} layouts our fundamental governance of data, consisting of public datasets and synthetic datasets through various data augmentation and simulation strategies, 
Section \ref{sec:agentic_sensing} introduces our approach to sense external knowledge from multiple sources,
Section \ref{sec:foundation_model} unveils the design of \letsfm{}, including its implementing framework, and detailed pre-training and forecasting performance, 
Section \ref{sec:multimodel_model} illustrates our trial model for multimodal learning,
Section \ref{sec:model_fusion} explains the model fusion work,
Section \ref{sec:adoption} showcases the performance of \leforecast{} in three real enterprise scenarios, 
Section \ref {sec:outlook} presents a discussion on deviations from existing TSFMs, along with our strategic blueprint, 
Section \ref{sec:conclusion} concludes the present delivery of \leforecast{} by the time of writing.

\clearpage
\section{Information Landscape: Time Series Data Governance}
\label{sec:data_goven}
\begin{figure}[h]
    \centering
    \includegraphics[width=0.8\linewidth]{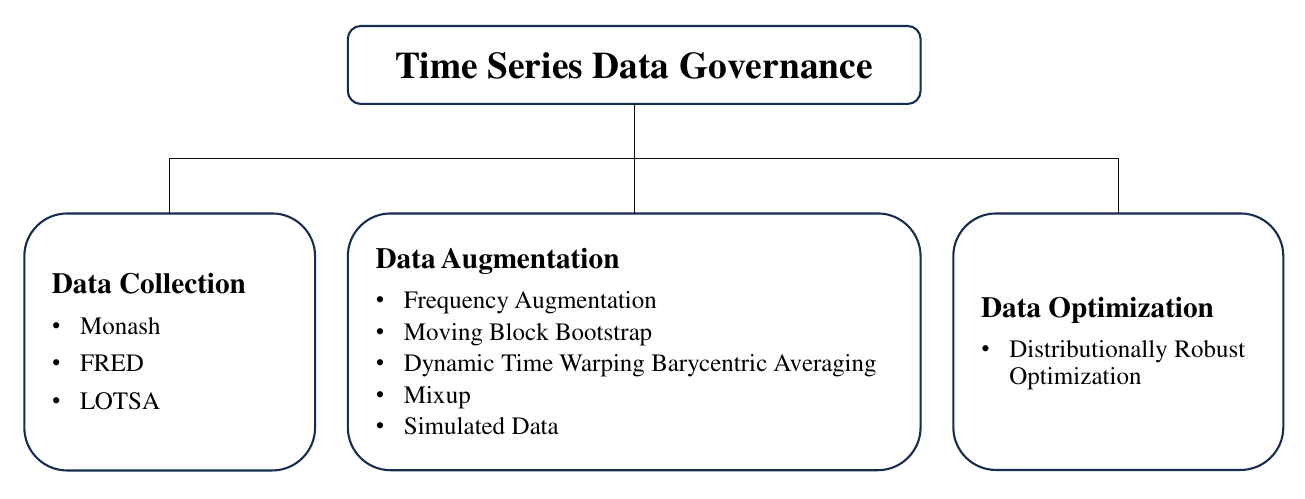}
    \caption{Framework of Time Series Data Governance}
    \label{fig:frame}
\end{figure}

Regarding time series data from multiple information sources, we design a data governance module to manage the quality and integrity of data for pre-training, monitoring, and validating time series forecasting capabilities \cite{Zha2023DatacentricAI}. To keep in line with existing work in the field of time series foundation models, data governance starts by collecting public time series data as long as we could access, then extends their breadth and depth via data augmentation and data optimization, thereby, enhances the generalization capability of \letsfm{} in pre-training. This section will brief the public datasets and then discuss our augmentation and optimization approaches on top of those datasets. The overall framework of time series data governance is shown in Figure \ref{fig:frame}.  


\subsection{Public Data Source}

We gather time series data from various sources, including the Monash Time Series Forecasting Repository \cite{godahewa2021monash}, Federal Reserve Economic Data (FRED) \cite{fred_all_data}, and the Large-scale Open Time Series Archive (LOTSA) \cite{Woo2024UnifiedTO}.

\begin{itemize}
    \item Monash Time Series Forecasting Repository is an open-access collection of time series datasets managed by Monash University. It includes a wide variety of real-world data from diverse domains, such as economics, finance, energy and healthcare. 

    \item FRED is an extensive online database maintained by the Federal Reserve Bank of St. Louis. It provides access to a wide range of economic data, including indicators, such as inflation, employment and GDP. Most of the data in FRED are at a coarse granularity, i.e. the time granularity ranges from monthly to yearly, making it ideal for macroeconomic analysis and trend exploration. 

    \item LOTSA is a collection of open time series datasets designed for pre-training large time series models. It consists of high-frequency and low-frequency data, which is essential and valuable for testing forecasting models and algorithms in various contexts. 
\end{itemize}

Currently, after dropping duplications while integrating these open datasets, data governance maintains time series data from different domains: Banking, Economics, Energy, Health, Nature, Sales, Tourism, Transport, Web and others, of which the time granularity ranges from minute, hour, daily, monthly to quarterly.

\subsection{Data Augmentation}
Apparently, the above mentioned open data is from real-world scenarios, implying several limitations for pre-training and testing models. First, data privacy and scarcity of time series impose a challenge in large foundation model pre-training, since the giant scale of time series is the fundamental factor for an efficient model. 
Additionally, time series from real-world reflects complex factors, which could vary from domain to domain and lack of basic time series patterns for a foundation model to learn general forecasting capability. 
Therefore, to further expand the breadth and depth of open data, we implement the following 5 strategies for data augmentation: 

\paragraph{Frequency Augmentation} High-frequency time series are aggregated into lower-frequency ones, such as converting minute-level data into daily-level data. This will enrich the diversity of time series patterns which have been demonstrated at various frequencies.

\begin{figure}[t]
    \centering
    \includegraphics[width=0.8\linewidth]{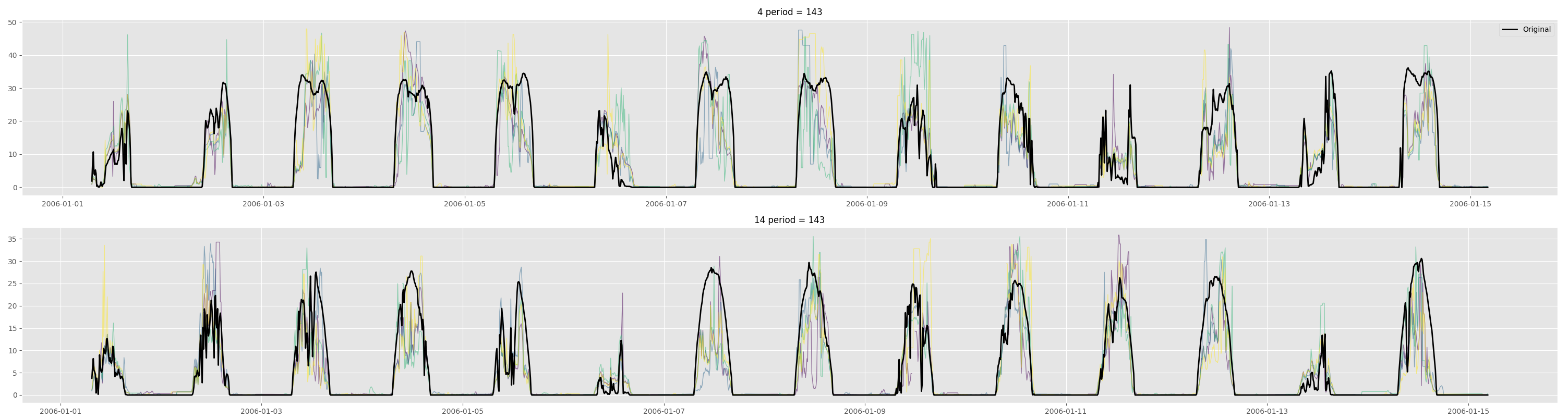}
    \caption{Example of MBB for Solar 10-Minute Dataset}
    \label{fig:solar}
\end{figure}

\begin{figure}[t]
    \centering
    \includegraphics[width=0.8\linewidth]{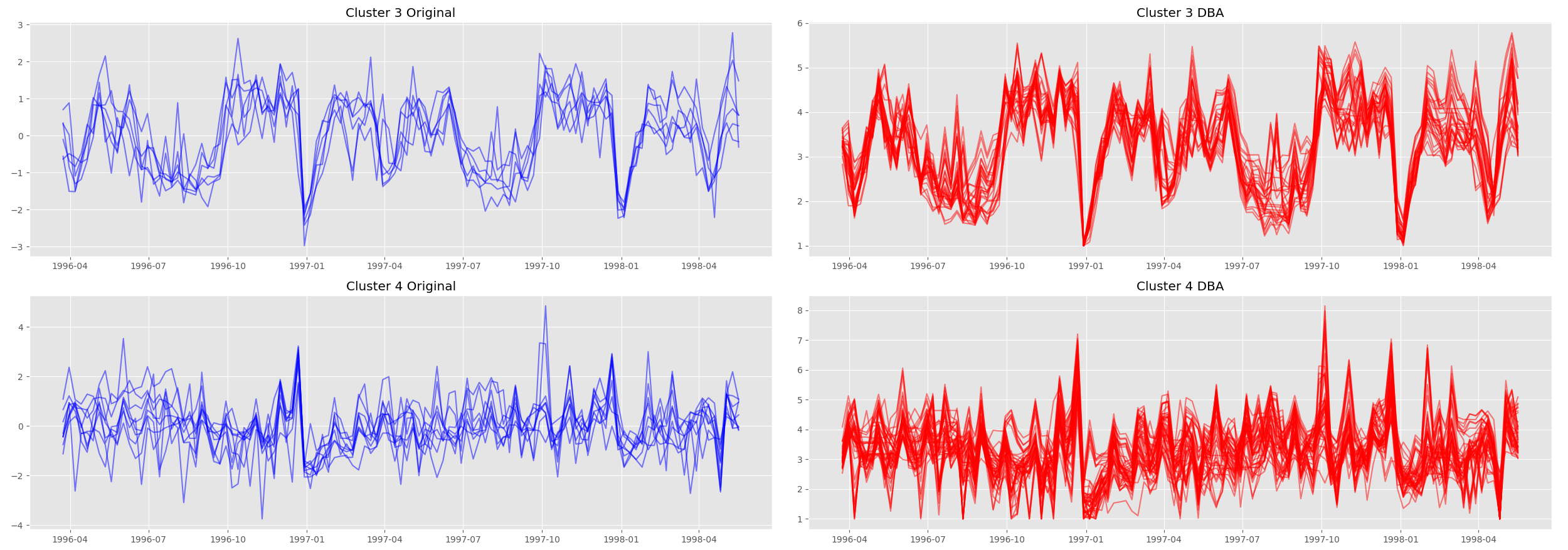}
    \caption{Example of DBA for NN5 Weekly Dataset}
    \label{fig:nn5}
\end{figure}

\paragraph{Moving Block Bootstrap (MBB)} This approach is a non-parametric approach for bootstrap resampling. It has been widely adopted in the bootstrap procedure on time series due to its advantages in capturing correlations along time series among overlapping blocks. We use MBB procedure to generate multiple synthetic versions of individual time series in the above collected datasets. First, we use locally estimated scatterplot smoothing (LOESS) to decompose the time series into three additive components: season, trend and residual (Seasonal-Trend Decomposition, STL). Then, bootstrapping the residual component into overlapping blocks. With these bootstrapped residual blocks, randomly resampling and concatenating blocks to the length of original series and gather in order before adding back the decomposed trend and seasonal components to generate a new bootstrap sample. By leveraging the original observations during bootstrapping, this approach ensures that the generated data preserve key statistical properties, including seasonality and trend, and captures a wide range of variations as well. Figure \ref{fig:solar} illustrates the effect of applying the MBB method to enhance two sequences from an original open dataset, Solar. The black lines represent the original sequences, whereas the colourful lines show the augmented variations.

\paragraph{Dynamic Time Warping Barycentric Averaging (DBA)}
Different from MBB that processes each series independently, DBA \cite{Petitjean2011AGA} iteratively uses dynamic time warping to align multiple time series and generates new synthetic samples by averaging them into an averaged series with minimum squared distance to all series. Therefore, the newly generated time series combine characteristics and reflect global characteristics of the given group of time series. In our practice, to maintain authenticity, we first use the k-shape \cite{Paparrizos2016SIGMOD} method to group time series into different clusters based on similarity, and then apply DBA within each cluster to generate new time series. Figure \ref{fig:nn5} illustrates the effect of applying the DBA method to enhance two clusters from the NN5 weekly dataset. The blue lines represent the original time series, while the red lines show the augmented versions. 

\paragraph{Mixup} It is a data augmentation principle, initiated from the computer vision domain, that constructs similar but different training examples with respect to the training distribution at minimum vicinal risk \cite{Zhang2017mixupBE}. In addition, it creates linear interpolations between pairs of samples, leading to smoother decision boundaries and improved model robustness and generalization. It is arguable that this method could also contribute to time series augmentation. 
In our deployment, we adopt the linear combination of randomly select multiple series to generate synthetic data, where individual weights of the selected time series are sampled from a Dirichlet distribution. 


\begin{figure}[t]
    \centering
    \includegraphics[width=0.8\linewidth]{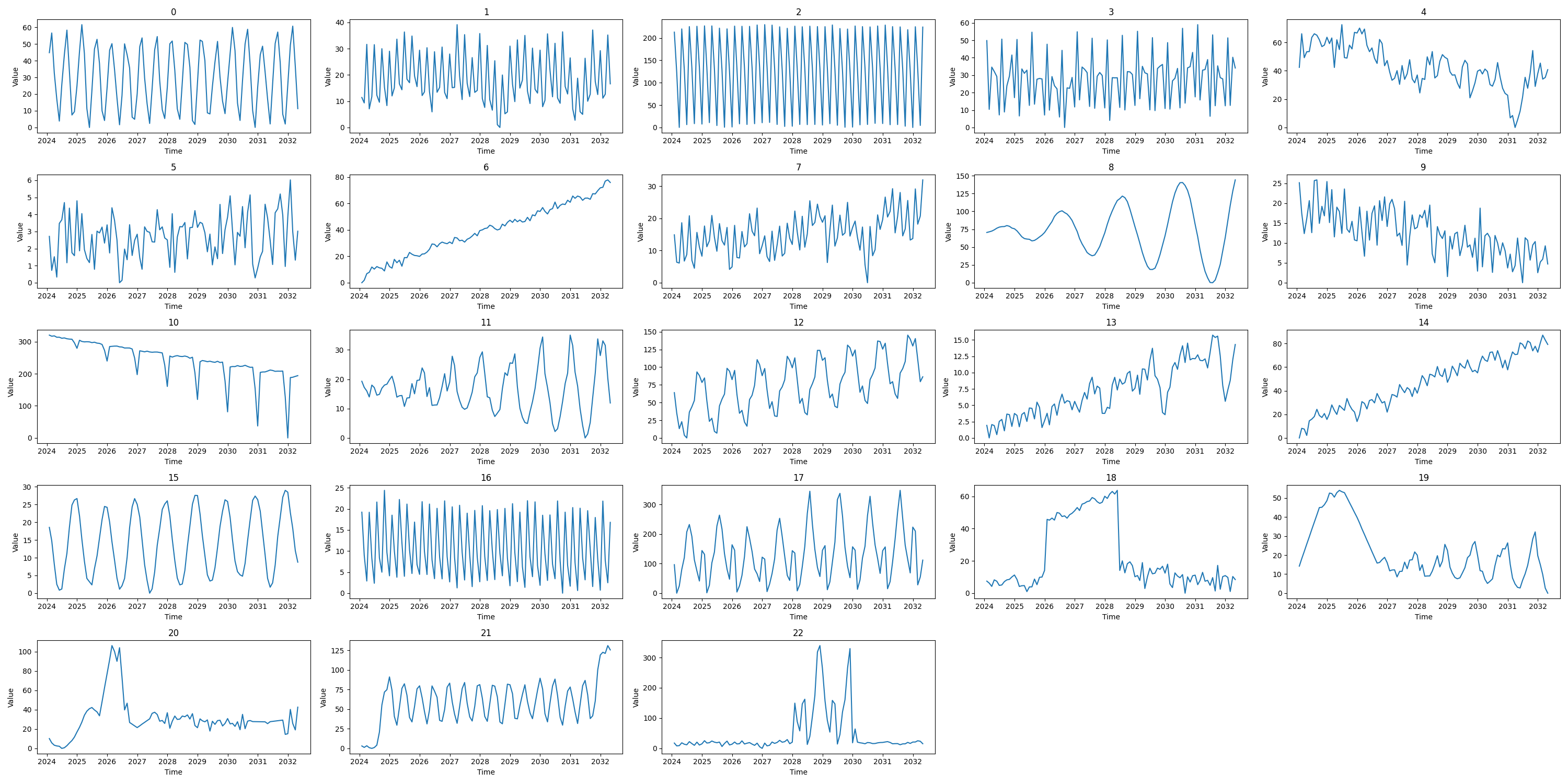}
    \caption{Simulated Time Series}
    \label{fig:skill_eg}
\end{figure}

\paragraph{Simulated Data}
In addition to real-world datasets, we create simulated time series that reflect a variety of characteristics, such as trend-dominant patterns, seasonal variations, noise influence, and state transitions. These simulated datasets shall model diverse real-world scenarios, hence, enhance the robustness of our foundation model. 
Specifically, we define time series as various compositions of three components: trend, seasonality and noise, in additive or multiplicative manner. Therefore, to generate simulated data, the first step in our process is to generate the three components with variations. We choose linear and exponential functions as base functions to construct trends, cosine functions and random periodic functions as seasonality construction alternatives, and Gaussian noise and red noise as random variations of noise component. Then, we permute the compositions of the above components to generate simulated data. For instance, composing an upward trend constructed by linear functions with a Gaussian noise samples in additive manner, results in a time series with an upward trend with random fluctuations. If the composition is conducted in the multiplicative manner, the series shall exhibit a different simulate characteristic, i.e. an upward trend with increasing noise variance instead. Similarly, combining a linear downward trend with a seasonality component produces a series that exhibits both a downward trend and periodic variations. By stitching together two series with different simulation characteristics, a time series can simulate state transitions. Figure \ref{fig:skill_eg} displays 23 examples of simulated time series, which replicate the diversity and complexity of real-world time series as closely as possible.

\subsection{Data Optimization}
\begin{figure}[t]
    \centering
    \includegraphics[width=0.8\linewidth]{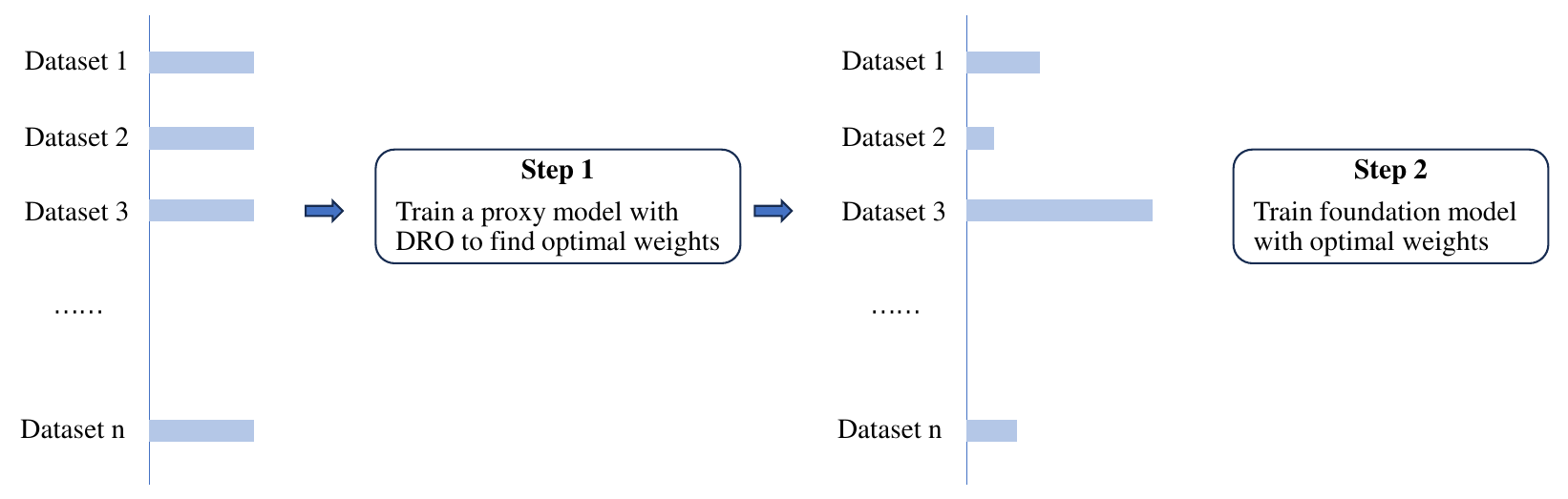}
    \caption{Optimizing Dataset Distribution for Time-Series Pre-Training}
    \label{fig:dro}
\end{figure}

It is widely acknowledged that the distribution of the pre-training data significantly affects the effectiveness of a foundation model. Determining an optimal data composition to balance generalization ability across domains is challenging. Within our operation, we leverage a distributionally robust optimization approach, DeReMi \cite{Xie2023NEURIPS} to address this issue. However, unlike natural language domains, time-series data from different domains do not exhibit significant structural differences. Therefore, instead of focusing on domain-specific characteristics, we optimize the weightings of datasets to balance their contribution during training. The process is illustrated in the Figure \ref{fig:dro}.  

We train a small reference model for every dataset as a benchmark of best training loss over individual datasets. Thus, we leverage the Group DRO optimizer, which calculates weighted cumulative loss across dataset and dynamically updates the weights based on the loss incurred in each dataset with respect to its benchmark loss. This process keeps adjusting the training objective, i.e. dataset weights, to minimize the relatively excess loss across datasets, contributing to an optimal objective for pre-training.

\subsection{Summary}
In summary, to establish a solid data foundation for time series foundation models, the data governance module collects diverse time series data from 13 different domains, ensuring a comprehensive representation of real-world patterns. It also applies five data augmentation strategies in constructing synthetic datasets, further expanding the collected data and enhancing model robustness. Additionally, we optimize the distribution of training data using a DRO-based approach, ensuring that the pre-trained model maintains balanced performance across all domains, mitigating domain shift issues. This approach significantly improves the model’s ability to generalize across various scenarios. In total, the data governance module maintains 22.3 billion data points for model training, with 13.7 billion sourced from real-world datasets, 0.6 billion generated through simulation and augmentation, and an additional 8 billion obtained via the mixup method. These extensive datasets not only improve the foundation model's accuracy and reliability but also increase its adaptability for a wide range of time series forecasting applications, making it a highly valuable tool for both industry and research.

\begin{table}[h!]
\caption{Pretraining Data Overview}
\centering
\begin{tabular}{|l|p{3cm}|p{3cm}|p{3cm}|}
\toprule
\textbf{} & Number of Datasets & Number of Data Entries & Number of Data Points \\ \midrule
Real-world Dataset & 160  & 2M  & 13.7B \\ 
Simulation and Augmentation & 30  & 0.4M & 0.6B  \\ 
Mixup & 6  & 3.9M  & 8B  \\ 
Total & 196  & 6.3M  & 22.3B  \\ 
\bottomrule
\end{tabular}
\end{table}

\clearpage
\section{Information Landscape: Information Mining}
\label{sec:agentic_sensing}
Knowledge and information enable the time series analyst to sense additional signals and trends beyond the historical data.
However, enormous experts spend a tremendous amount of time working as a group to survey, analyse and discuss market status, trend, or expectations to adapt general forecasts to complex knowledge, 
we intend to fulfil the gap between massy external information and systematic analysis, where Information Mining is initiated. 
Information Mining is designed to adopt cutting-edge research in the field of large language models (LLMs) to format and inject such external information into time series forecasts. 
LLMs are known as an intelligent tool to answer queries while autonomously interacting with various inputs following specific instructions. 
However, complex instructions introduce a significant degree of uncertainty into generated answers, especially for multiple various data sources. 

\begin{figure}[h]
    \centering
    \includegraphics[width=1.0\linewidth]{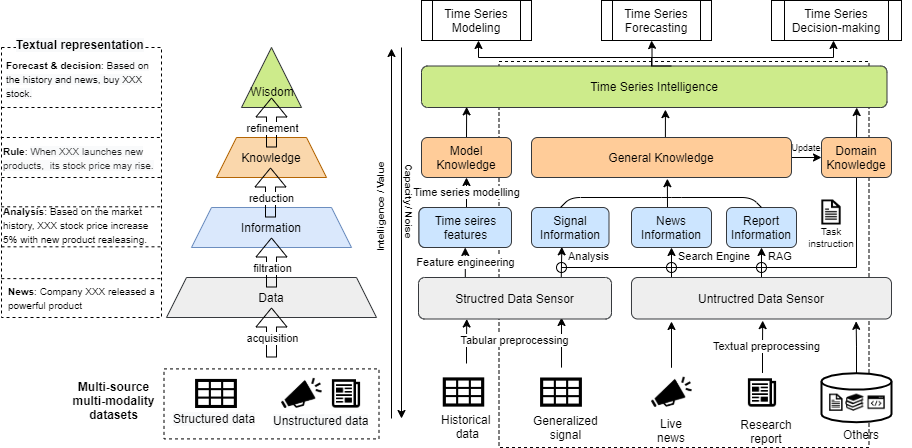}
    \caption{Information Mining Framework}
    \label{fig:Mkss_overview}
\end{figure}

To proactively analyse associated data about forecasting targets, such as market status, before making forecasts, Information Mining adopts a multi-sensor framework to conduct analysis across multiple sources nominated by business experts, described in the right dotted part in Fig \ref{fig:Mkss_overview}. 
In this section, we discuss designated sensors and an analyst building upon LLMs in accordance with two types of external information sources for information mining in our practice.




\subsection{Information Sources}
According to the Data-Information-Knowledge-Wisdom (DIKW) hierarchy theory in  \cite{rowley2007wisdom}, the value gradually increases with the hierarchy level rising, while noise gradually decreases by filtration, deduction, and refinement.
The detailed textual representation samples and hierarchy is shown in the left part in Fig \ref{fig:Mkss_overview}.
Based on our practical experience in the past decade, we first categorise external information sources into Structured Data and Unstructured Data, regarding presentation structure.  

\begin{itemize}
    \item \textbf{Structured Data}: 
    Historical data is the essential structured data to analyse the time series.
    In addition, business datasets offer domain information in form of structured data, such as generalized signal and historical data from the adjacent channel. 
    To absorb such kinds of information in accordance with forecasting targets and domains, we categories them first. 
    Taking PC shipment forecasting as an example, structured data including manufacturing plan, sales outlook, and customer activations are all labelled as relevant information. 
    \item \textbf{Unstructured Data}: 
    Obviously, there is more information contained beyond structured data.
    Live news offers real-time information about business supplements, markets and customers which implicate potential impact on business. For instance, breaking news on extreme weathers, strikes and trading restrictions could disrupt supply chain for various durations. 
    To collect live news according to their relevance to a forecasting target within a specific domain, we select the timestamp, source, title as the keywords and retrieve the content from the search engine API.
    


    Research report is another pivotal source of external information leveraged in business. 
    Business experts have been expressing that business decisions or strategies would be made in consideration of reported trends, potentials, etc from reliable parties. 
    Therefore, in alignment of real operations, we collect relevant research reports with guaranteed updates for each forecasting targets.
    Basically, consulting firms, securities companies, rating agencies, government agencies, industry associations provide the business report on a regular basis.

    Other unstructured data including social media sentiment, surveys, blogs, and books is all promising to provide external information to help the down-stream analysis as well. 

\end{itemize}

Note that the above types of information sources complement each other.
Business report data ensures higher relevance but cannot guarantee the all-time coverage, where live news data provide timely knowledge with highly redundant.
Generalized signal consider both relevance and coverage of knowledge, although appropriate supplement data is not always available for each target variable.
In addition, these data sources cover multiple frequencies, with news data providing daily precision and reports data ranging from weekly to monthly and yearly intervals.

\subsection{Data Sensor}

LLMs are well-recognised as an intelligent generative tool to answer queries following specific instructions. 
It is capable of understanding long-term trends and timely changes of query series. 
However, complex sensing instructions introduce a significant degree of uncertainty into generated answers. 
Information Mining uses a multi-sensor framework to analyse both long-term and short-term knowledge acquired from above knowledge sources and offer timer series foundation modal an either positive or negative sense on status and trends of forecasting targets with a confidence index. 
Specifically, as Figure \ref{fig:Mkss_overview} shows, Information Mining consists of multiple sensors and an Analyst.
In this section, we introduce Signal Sensor, News Sensor, and Report Sensor as examples.
\paragraph{Structured Data Sensor}
Signal Sensor illuminates the brain of business experts, i.e. analysing structured data (i.e. tabular data) with respect to linguistic human knowledge, and inferring future trend reflected by the data in consistence with prior knowledge. 
For this purpose, Signal Sensor reshapes data structure and converts human knowledge into clear instructions for data analysis, including rules to analyse data correlation, link numerical changes based on business contexts, infer possible reasons or outcomes from statistics, etc. 
Thereby, Signal Sensor derives informed summary from original data including either an positive or negative sense on the data trend with a confidence index and brief reasons. 
An example prompt designed for Signal Sensor can be found in Appendix \ref{sec:appendix-b-data}.

\paragraph{Unstructured Data Sensor}
News Sensor utilises a search engine \cite{News2024Tavily} built specifically to work with LLMs to automatically collect real-time and accurate news from WWW at a sufficient speed, and also loops human in website verification in terms of reliability and relatedness within a specific business domain, before acquiring a summary from LLM with a designated prompt. 
\begin{figure}[t]
    \centering
    \includegraphics[width=\linewidth]{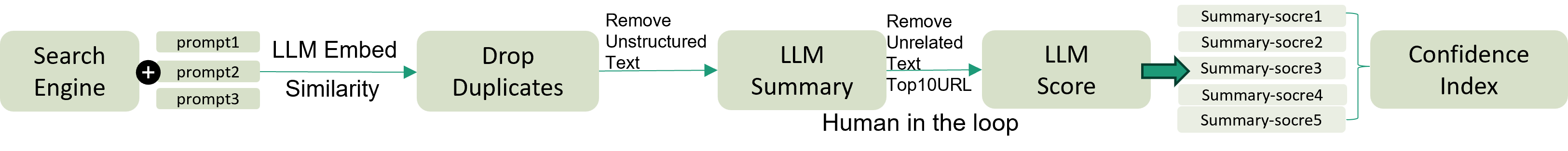}
    \caption{Workflow of News Sensor}
    \label{fig:news_agent}
\end{figure}
Figure \ref{fig:news_agent} illustrates the whole workflow of News Sensor. 
Starting at searching accessible and primary websites related to given task instruction, the sensor invokes LLMs to perceive collected webpages and drop redundant ones as well as unrelated HTML formats, before generating a brief and precise summary of each webpage. 
The summaries will be emphasized and filtered by human experts, and then scored by LLMs. 
The last action is to invoke LLMs, with previous highly summarised webpages and a well-tested prompt, to classify viral market status as either positive or negative with a confidence index, and generate a reason for its decision. 
An example prompt used by News Sensor can be found in Appendix \ref{sec:appendix-b-websearch}.

Report Sensor is in charge of reading business reports, and then extracting insights regarding trends of specific markets. 
To fulfil such a task, Report Sensor overcomes two issues: retrieve most related materials and guide LLMs to understand the time sensitivity of reports. 
The first challenging issue is the report format, we use the latest retrieval-augmented generation (RAG) techniques to embed the text, the images, and tabular data into one representation.
Then, the freestyle in report writing, i.e. some topic could be mentioned but not as the leading cast in a scenario affect the performance. 
Noticing the consistent content structure in given reports, headline, bullet-points and then details, Report Sensor implements a hierarchical approach to first narrow down discussions focused on a specific market, and then retrieve related perspectives of the market in details. 
At last, Report Sensor has a sense of market status and summarise it as either a positive or negative tag with a confidence index and brief reasons. 
An example prompt designed for such web search can be found in Append \ref{sec:appendix-b-report}. 

\subsection{Knowledge Integration} 
We design an Analyst module, that gathers all of the analysis from above individual sensors and fuses those analysis into a comprehensive analytical report on given tasks.
Specifically, the Analyst collects the analysis from different sources and make summaries respectively.
The final summary for each forecasting target is then produced by LLM to evaluate the received multi-source knowledge. 
An example of Analyst prompt can be found in Appendix \ref{sec:appendix-b-report}. 

The Analyst is also needed to output the quantized trend prediction for the target variable.
Firstly, we quantify the analysis results of each information source, with 100 points for positive ones and 0 points for negative ones. 
Then, we calculate the overall weighted average score, where the weight can be the analysis confidence itself.
Note that such weight is not the proper calculation for the accurate description.
For instance, for the analysis of a computer market, instead of putting the trends derived from computer shipment data and the news about newly launched computers at the same weight, forecasting the demand of computers shall value the latter information source more. 
We propose a dynamic weight learner to obtain the influence weight of specific information source by training the forecasting target with the timestamp-aligned information material. 
Finally, we obtained the trend prediction score for the future by the summation calculation of future trend and each influence weight. 
It is a score ranging from 0 to 100. 
The closer the score is to 100, the more positive trend it indicates that the target variable in the future will be based on the multi-source information.
The higher the score is, the more confident it is in the judgment that the target variable will increase in the future.
Obviously, the closer the score is to 0, the more negative it implies.

\subsection{Summary} 
Information Mining enables \letsfm{} to perceive external environment via analysing information and data from multiple sources, including live news, business report and related data, thereby, to build a comprehensive understanding of contextual knowledge of universal time series analysis. 
In our vision, potential future enhancements of Information Mining are the followings:

\textbf{Long-term trend analysis from the sensing material.}
The Information Mining system is capable of analysing the potential trend in close future, whereas the time window to forecast in real business scenarios can be a far future and the probability that the hallucination of a LLM creates is greater. 
Autoregressive analysis is promising to derive long-term trends instead of hallucinations. 

\textbf{Task-oriented fine-tune for the News Sensor.}
The performance of News Sensor depends on the intelligence level of LLMs to understand, identify and summarise information from massive websites. Fine-tuning is one of approaches to enhance LLM intelligence and sensitivity in specific areas or topics, yielding better searching results and correlation analysis over valuable news and contents. 

\textbf{Introduce the debate mechanism}
Debate mechanism is promising to enhance the reasoning capability of LLM. 
A Reasoner module provide inference of future trend based on the collect multi-source information.
Meantime, a Refuter module can be designed to rebut the inference from the Reasoner using the historical evidence.
Besides, adaptive RAG, layered memory, output normalization are potential approaches to improve the Information Mining work.

\clearpage
\section{Modelling Pillar: Time Series Foundation Model}
\label{sec:foundation_model}


Foundation models refer to advanced machine learning paradigms that are pretrained on massive and diverse data and presumably adaptive for various downstream tasks. Numerous domains have seen rapid development of foundation models in recent years, yielding great potential for business acceleration in the future. In the field of time series forecasting, however, there are comparatively limited studies of foundation models pre-trained on extensive time series for forecasting tasks. We intend to build a time series foundation model, which will reduce the need for extensive customisation and model tuning across domains, and maintain generalisability, reliability and scalability at the same time. This leads us to \letsfm{}, the Lenovo Time-Series Foundation Model with a scalable transformer backbone, leveraging extensive time series from Data Governance in Section \ref{sec:data_goven}. In this section, we discuss the structure and pretraining of \letsfm{}, which contributes to its robust and generalised forecasting performance in comparison to peer foundation models on diverse datasets.

\subsection{Model Structure and Pretraining}
\begin{figure}[h]
    \centering
    \includegraphics[width=0.8\linewidth]{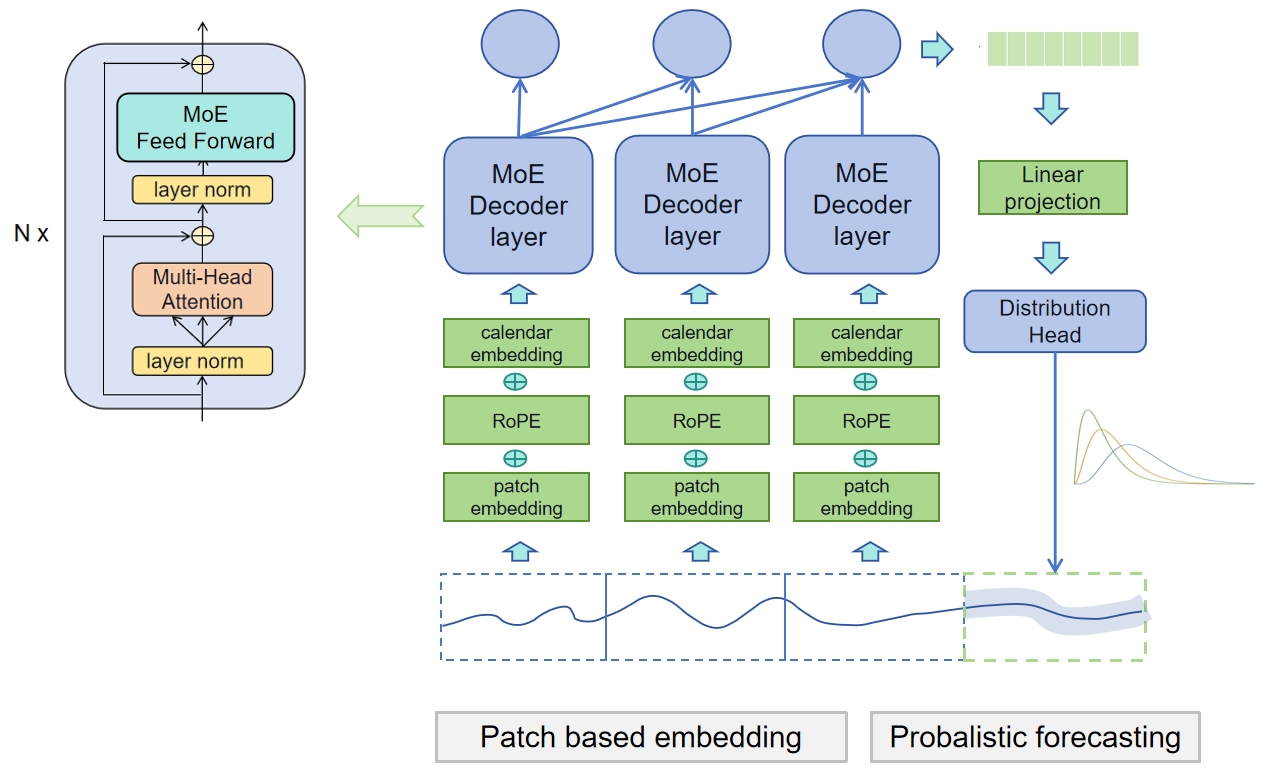}
    \caption{Architecture of Le-TSFM}
    \label{fig:Le-TSFM model structure}
\end{figure}

Lenovo Time-Series Foundation Model (\letsfm{}) employs a GPT architecture and is pre-trained using large-scale data for time series forecasting tasks.
As Figure \ref{fig:Le-TSFM model structure} illustrates, \letsfm{} features a MoE, decoder-only Transformer architecture, which comprises 6 layers, a model dimension of 1024, 4 experts in a mixture-of-experts (MoE) framework, 8 attention heads and a total of 150 million parameters hereof. Apart from the classic transformer structure, \letsfm{} has three distinct designs for time series forecasting task: multi-scale patch tokenization, patchwise multi-step probabilistic forecasting and joint embedding for time series patch, calendar information and sample frequency.

To aggregate local semantic information around individual data points in time series, \letsfm{} first segments time series into multi-scale patches, which will serve as its input tokens. Previous studies have proved that patching time series facilitates context-sensitive perception in transformer models and optimises the processing efficiency of long time series \cite{nie2022time}.  On top of time series patches, \letsfm{} explicitly takes timestamps and frequency of time series as input as well, leveraging separate linear layers for patch embedding, calendar-related information embedding and sampled frequency embedding, as Figure \ref{fig:Le-TSFM model structure} shows.

In pre-training phase, \letsfm{} leverages the dataset of 22 billion data points maintained by Data Governance (Section \ref{sec:data_goven}). To address the imbalance in domain representation and sampling frequency in data, we conduct the pretraining with the paradigm of distributionally robust optimization \cite{sagawa2019distributionally}. In particular, \letsfm{} dynamically adjusts weights of individual domains based on excessive loss between its pre-training and reference loss, which enables stable generalization across diverse scenarios with balanced converging speed of training losses from multi-domains.

Moreover, \letsfm{} casts time series forecasting task as patchwise probabilistic forecasting, i.e. forecasting
distribution of the next few time steps via a parametric distribution estimation. Hence, it adopts negative log-likelihood loss in pretraining via forecasting the parameters of a Student-T distribution, which is equivalent to providing probabilistic predictions for multiple future time steps simultaneously. Additionally, when the forecast horizon exceeds the patch length in inference phase, \letsfm{} generates forecasts recursively for the specified length.

\subsection{Forecasting Performance and Generalization Capability}

For pretrained foundation models, it is widely accepted that performing accurate zero-shot forecasting is a fundamental capability of robust and generalised models, especially delivering forecasts across diverse data distribution and sample size. The capability of zero-shot forecasting is particularly versatile and beneficial for enterprise where numerous business scenarios lack of adequate and reliable data. Therefore, to validate the potential of \letsfm{} to serve enterprise systems as the backbone to produce reliable, accurate and stable forecasts, we evaluate the capability of \letsfm{} in zero-shot forecasting. 

\begin{figure}[ht]
    \centering
    \begin{subfigure}[b]{\textwidth}
        \includegraphics[width=\textwidth]{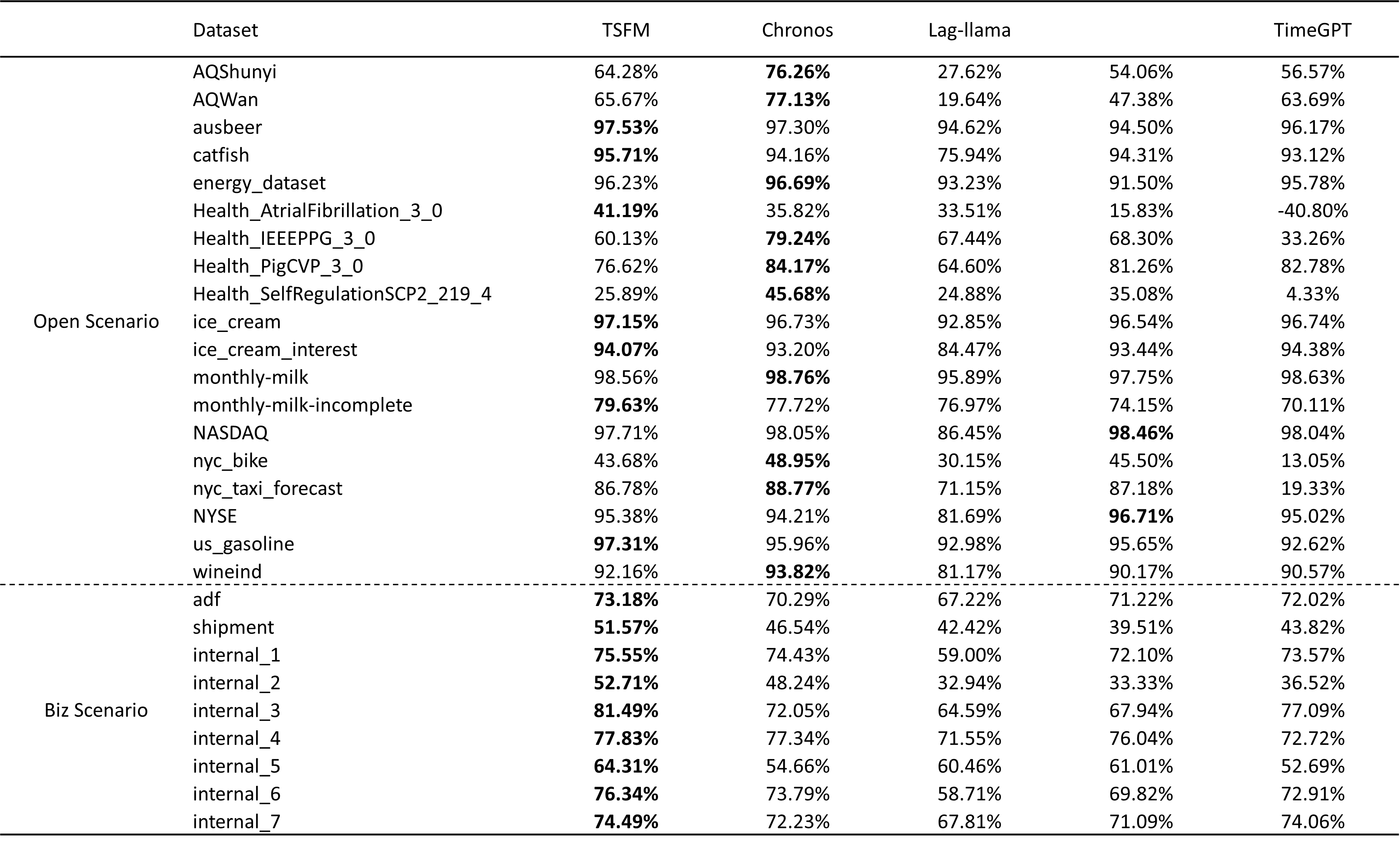} 
    \end{subfigure}  
    \begin{subfigure}[b]{\textwidth}
        \includegraphics[width=\textwidth]{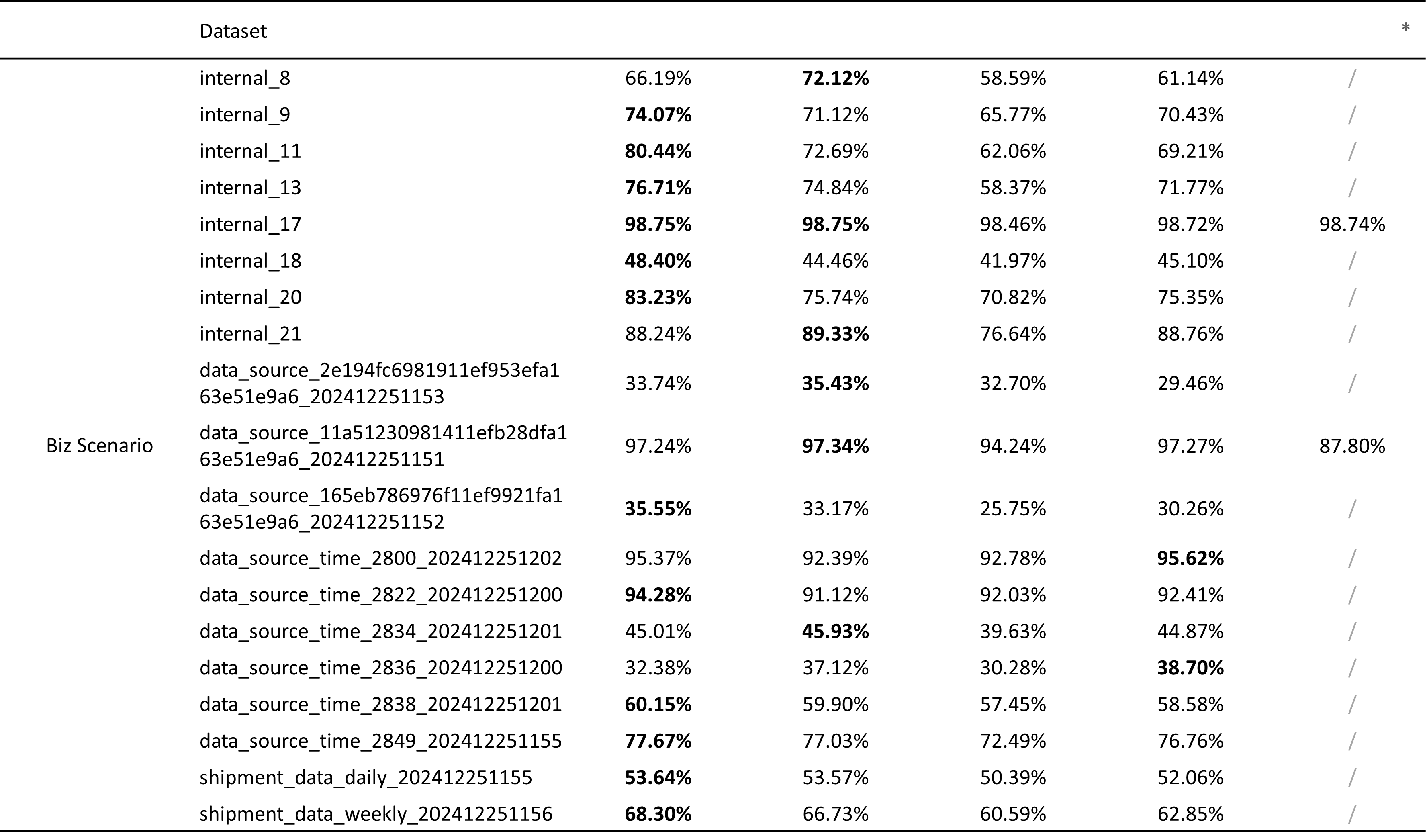} 
    \end{subfigure}
    
    \caption{Zero-shot forecasting benchmark across various domains of Lenovo's internal datasets (Biz Scenario) and open source datasets (Open Scenario). Metric shown in table is 1-MAPE. Both internal and external datasets are carefully curated to exclude those contained in the pretrain dataset pool of \letsfm{} or peer TSFMs}
    \label{fig:Le-TSFM_benchmark}
\end{figure}

To be specific, we conduct extensive experiments on 19 external datasets (Open Scenarios) from various domains and 34 internal datasets collected from various Lenovo business scenarios. Figure \ref{fig:Le-TSFM_benchmark} presents the detailed experiment results. The external datasets span multiple domains, including industry, finance, retail, transportation, healthcare and meteorology, representing the diversity of real-world time series data. The internal datasets encompass diverse scenarios within the full workflow of Lenovo's intelligent manufacturing and supply chains, including product demand, spare parts allocation, retail prices, logistics capacity, production sensor data, and failure rates of components. For a fair benchmark test, we rigorously curated the above evaluation datasets to eliminate overlaps with the pre-training data of \letsfm{} and peers: Chronos \cite{ansari2024chronos}, Moirai\cite{woo2024unified}, Lag-Llama \cite{rasul2023lag}, to avoid potential information leakage.

\begin{figure}[ht]
    \centering
    \begin{subfigure}[b]{0.7\textwidth}
        \includegraphics[width=\textwidth]{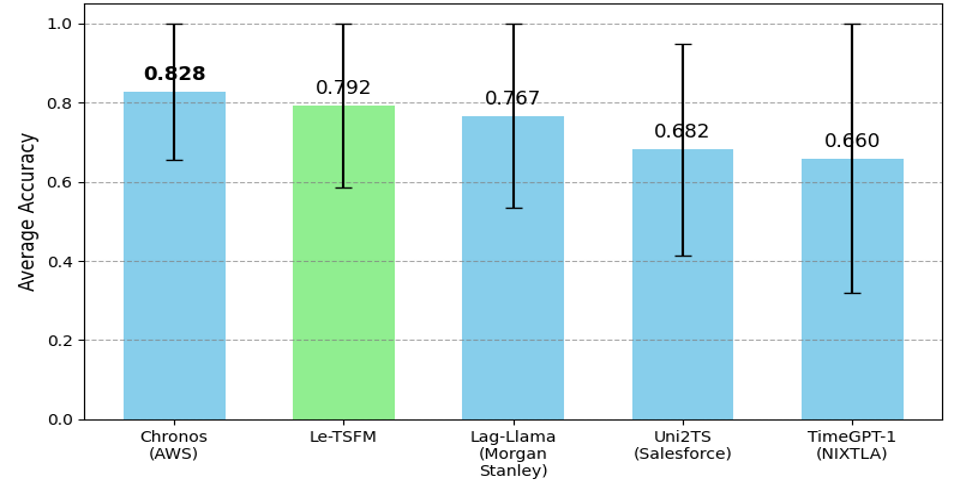} 
        \caption{Average Performance on Datasets of Public Scenarios}
        \label{fig:subfig1}
    \end{subfigure}
    
    \vspace{1cm} 
    
    \begin{subfigure}[b]{0.7\textwidth}
        \includegraphics[width=\textwidth]{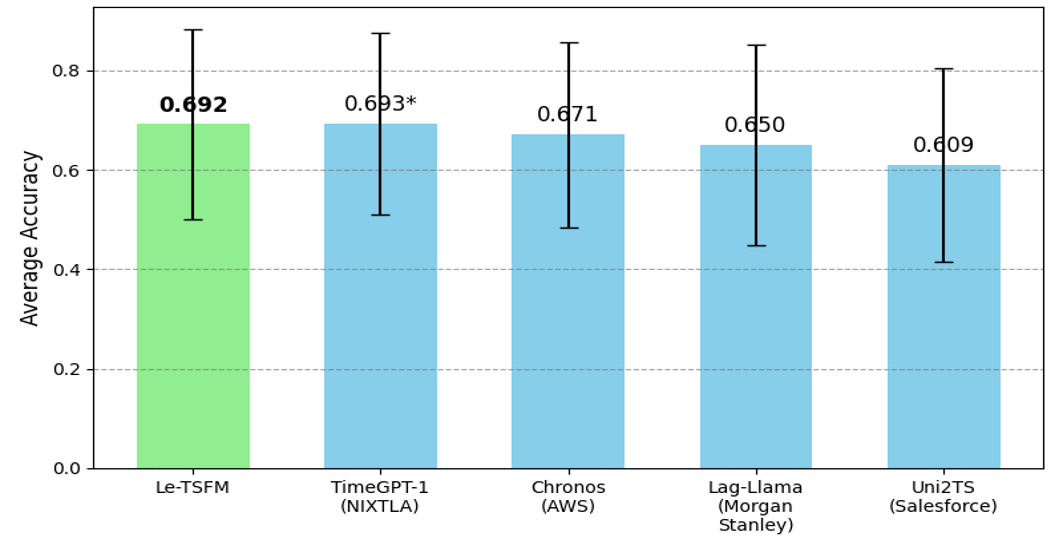} 
        \caption{Average Performance on Datasets of Business Scenarios}
        \label{fig:subfig2}
    \end{subfigure}
    
    \caption{Average Zero-shot Performance of TSFMs}
    \label{fig:Le-TSFM_benchmark_avg}
\end{figure}

As indicated in Figure \ref{fig:Le-TSFM_benchmark} and \ref{fig:Le-TSFM_benchmark_avg}, \letsfm{} has the edge in zero-shot forecasting over its peers or achieves comparable performance. 
For the experiments with internal datasets, \letsfm{} performs superior forecasting accuracy in 21 out of 34 datasets with an optimum 10\% uplift from the second best forecasts. Additionally, \letsfm{} surpasses its peers in 7 out of 19 external domains by around 6\% forecasting accuracy at the best. Overall, \letsfm{} exhibits remarkable stability across diverse datasets and significantly outperforms its peers in terms of consistency. It is convincing that such significant robustness is one of the beneficial effects of the domain-aware data augmentation strategies to mitigate data imbalance along with the incorporation of distributional robustness optimization techniques during pretraining. 

\subsection{Summary}
We have demonstrated the methodological value of foundational models in time series forecasting tasks. By leveraging a GPT-style scalable architecture and incorporating the basic pretraining paradigm of next patch forecasting, we enabled the model to achieve generalizable and transferable prediction capabilities. Beyond model architecture, the enhancement of the pretraining datasets is crucial to mitigating the severe imbalances in domain coverage, sampling frequency, and data patterns inherent in public time series datasets. Additionally, to further strengthen the model's adaptability across varying data distributions, we adopted a Distributional Robust Optimization (DRO) framework during the pre-training process. Empirical evaluations using internal enterprise datasets across multiple domains and public benchmarks confirmed the proposed Le-TSFM’s effectiveness and reliability.
\clearpage

\section{Modelling Pillar: Multimodal Model}
\label{sec:multimodel_model}

In reality, there are complex linguistic contexts related to time series, such as descriptions, expectations as well as interpretations, which provide wealth and valuable knowledge to understand underlying patterns of time series yielding better forecasting performance. Injecting semantic knowledge into time series forecasting has been a well-established as an effective but challenging approach to improve time series forecasting performance. 

We propose a multimodal forecasting model, Dual-Forecaster, that learns from historical time series and their associated semantic descriptions, and then forecasts future horizon accompanied adaptation to semantic expectations, to enhance the forecasting efficiency and adaptiveness of \letsfm{}. Currently, the comprehensive introductions of model implementations and evaluations have been submitted for review. This section will only brief the concept of our multimodal model and illustrate its performance with a test case. The comprehensive version will be included shortly. 

\subsection{Multimodal Forecasting}

Given a dataset of \textit{N} numerical time series and their corresponding textual series, including descriptively historical texts which can be used to augment the model's capacity to learn the relationships between different time series, and predictive texts that can provide additional insights to assist the model in perceiving and dynamically adapting to event-driven time series distribution drift. The goal is to maximize the log-likelihood of the predicted distribution obtained from the distribution parameters learned by the model based on historical time series data and its corresponding descriptive textual information and predictive textual insights.

We design two branches: a textual branch and a temporal branch, to encode textual descriptions and time series, respectively, and two modality interaction modules: historical-oriented modality interaction module and future-oriented modality interaction module, for cross-modal alignment of encoded textural representations and time series representations within look-back time window and forecasting horizon, respectively. Furthermore, we compose pre-training loss with both historical text-time series alignment loss and time series forecasting loss, leading our model to understand both textual descriptions and time series at the same pace. 

\subsection{Forecasting Efficiency}

Extensive experiments on the synthetic dataset and four captioned-public datasets have validated our model's efficacy, with comparative analyses against six existing state-of-the-art models covering single- and multi-modal baselines \footnote{Single-modal Baseline Models: DLinear \cite{zeng2023transformers}, FITS \cite{xu2023fits}, PatchTST \cite{nie2022time}, iTransformer \cite{liu2023itransformer}. Multi-modal Baseline Models: Time-LLM \cite{jin2023time}, MM-TSFlib \cite{liu2024time}}. Experimental results show that our model consistently outperforms other models on all datasets. Moreover, ablation studies emphasize that the performance enhancement is attributed to the additional information provided by text.

\begin{figure}[t]
    \centering
    \includegraphics[width=\linewidth]{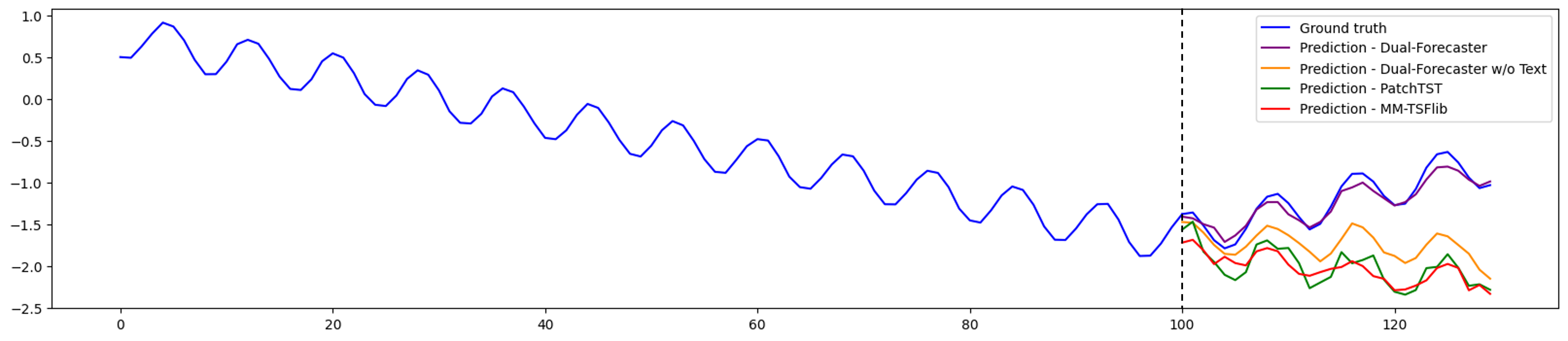}
    \caption{Visualisation of an example from synthetic dataset under the input-100-predict-30 settings.}
    \label{fig:syn_dataset_vis}
\end{figure}

Figure \ref{fig:syn_dataset_vis} shows an example of a simulated time series, which exhibits a transition from downtrend in look back window (from timestamp of 0 to 100) to uptrend in forecasting horizon (from timestamp of 100 to 120), simultaneously keeping the seasonality unchanged. In this visualization, both PatchTST and MM-TSFlib maintain the state observed in the \textit{look back window}, indicating an inability to adapt to state transitions. Notably, in the absence of textual information input, Dual-Forecaster also tends to provide conservative forecasts akin to these two methods. In other words, the generated forecasts are extensions of time series patterns observed in the historical time series. Conversely, with the integration of textual insights, Dual-Forecaster can adaptively perceive potential future state transitions, thereby delivering more reasonable forecasts. This underscores the substantial benefits of incorporating textual data for time series forecasting.

\subsection{Summary}
Regarding multimodal information sources, we design an innovative multimodal time series model aiming at generating more reasonable forecasts. It is augmented by rich, descriptively historical textual information and predictive textual insights, all supported by advanced multimodal comprehension capability. To enhance its multimodal comprehension capability, we craft multiple cross-modality alignment techniques, including historical text-time series contrastive loss, history-oriented modality interaction module, and future-oriented modality interaction module. Based on our current experiments on synthetic dataset and captioned-public datasets, our multimodal model deliveries effective forecasting and superior ability in integrating textual data for time series forecasting. 

\clearpage
\section{Modelling Pillar: Hybrid Model Fusion}
\label{sec:model_fusion}


Due to the variety of data sources, model architectures, and training recipes, time series forecasting models each have their unique strengths. Model fusion, which belongs to the "learning from model" paradigm rather than the traditional 'learning from data' paradigm, allows for the integration of multiple models into a single, cohesive model or framework \cite{li2023deepmodelfusionsurvey,tang2024fusionbenchcomprehensivebenchmarkdeep}. It leverages complementary strengths of various models to obtain better generalization abilities and prediction performance \cite{jin2023datalessknowledgefusionmerging,wang2024fusingmodelscomplementaryexpertise}. 
To fuse various forecasting models developed for time series, this section presents our framework for time series model fusion, composed by a model pool, model profiling and two different fusion approaches regrading various of model architectures to leverages the capability of various types of models. Specifically, we presents two approaches for heterogeneous model fusion: router-base fusion network and coordination of large and small models, which are still at experimental stage and will be released soon.

\subsection{Model Fusion Framework}
\begin{figure}[ht]
    \centering
    \includegraphics[width=1\linewidth]{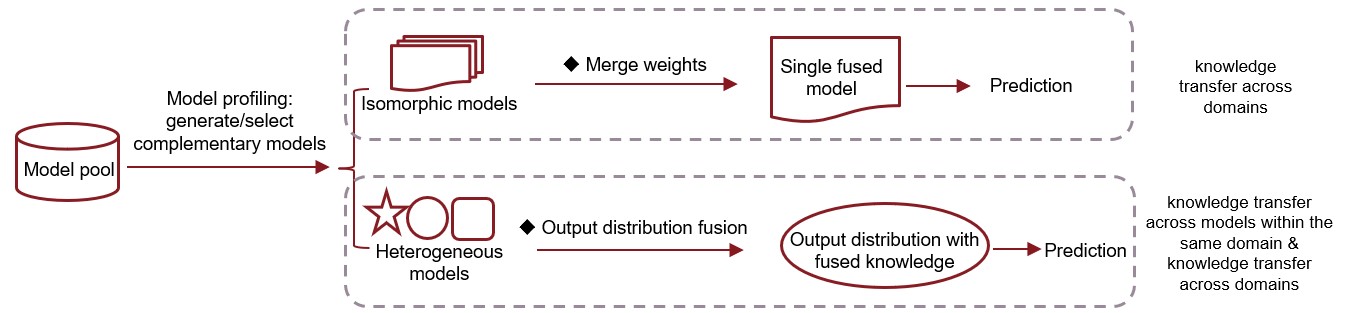}
    \caption{Model Fusion Framework}
    \label{fig:model_fusion_framework}
\end{figure}

The fusion framework for time series models, as Figure \ref{fig:model_fusion_framework} shows, consists of a model pool, model profiling, fusion of isomorphic models and fusion of heterogeneous models. Model pool stores fundamental models to be leveraged by fusion process. Without any constraints on model types, it encompasses time series foundation models (TSFMs, e.g., Le-TSFM, Chronos \cite{ansari2024chronos} and Moirai \cite{woo2024unified}), small-scale deep learning models (e.g., Transformer-based networks \cite{nie2022time}, recurrent neural networks \cite{lai2018modelinglongshorttermtemporal}, temporal convolution networks \cite{bai2018empirical,liu2022scinettimeseriesmodeling}), machine learning models (e.g. GBDT-based, XGB-based tree models), and statistical models (e.g. ARIMA \cite{7046047}).
The second module, model profiling, is a unified and scalable pipeline for model evaluation and model portrait. It will identify complementary models to enhance recovery of the ground truth best model for an input \cite{wang2024fusingmodelscomplementaryexpertise}. 

As for the third and fourth modules, fused models can be isomorphic or heterogeneous, thus their fusing strategies can be really different \cite{li2023deepmodelfusionsurvey}. In case of isomorphic model fusion, the models to be fused can be TSFMs or small-scale deep learning models. Since their architectures are identical, the fusion focuses on merging model parameters \cite{jin2023datalessknowledgefusionmerging,ilharco2023editingmodelstaskarithmetic,matena2022mergingmodelsfisherweightedaveraging,tang2024mergingmultitaskmodelsweightensembling}. Isomorphic model fusion often requires little or no training. Furthermore, it uses just a single model to achieve performance comparable to that of multiple models trained or fine-tuned in different domains, effectively facilitating knowledge transfer across various domains. In contrast, heterogeneous model fusion deals with models that have significantly different model architectures. In such scenarios, the outputs (or output distributions) of each model are fused rather than their parameters \cite{huang2024ensemblelearningheterogeneouslarge,wan2024knowledgefusionlargelanguage}. A common strategy is to train a router that determines the weights when integrating outputs from each model \cite{wang2024fusingmodelscomplementaryexpertise,lu2023routingexpertefficientrewardguided}. Heterogeneous model fusion can usually obtain better performance than each model to be fused, however, it requires sufficient training. In addition, this approach is more versatile, as it is not only applicable for knowledge transfer across different domains but also for knowledge transfer across models within the same domain.

Therefore, this fusion framework can fuse knowledge from multiple domains while also leverage universal and unique strengths of different time series models. In addition, it maintains flexibility to incorporate new models and supports iterative training, thereby enabling the platform to learn continuously. In the subsequent sections, we will show some experiments and observations conducted within this framework, focusing primarily on more general fusion strategy - fusion of heterogeneous models. Although the experiments are not exhaustive, they provide valuable insights and directions for future exploration.



\subsection{Router-Based Network to Fuse TSFMs}

As the community of time-series foundation models continues to grow, an increasing number of models are being open-sourced. These TSFMs have their own unique strengths due to differences in pre-training datasets/strategies and model architectures. Therefore, inspired by the "learning from models" paradigm, it is worthwhile to fuse these TSFMs to exploit the complementary potential \cite{jiang2023llmblenderensemblinglargelanguage,huang2024ensemblelearningheterogeneouslarge,lu2023routingexpertefficientrewardguided}. This approach does not require cumbersome processes of data collection, model design, and pre-training, potentially yielding superior performance compared to each TSFM to be fused. 

\begin{figure}[t]
    \centering
    \includegraphics[width=0.4\linewidth]{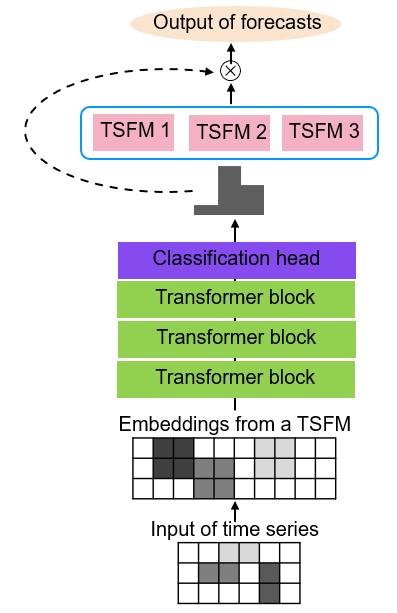}
    \caption{Concept of Router-based Network to fuse different TSFMs.}
    \label{fig:route_based_fusion}
\end{figure}

A router network is trained to automatically assign appropriate weights to each TSFM's output, thereby fusing different TSFMs into a unified prediction framework. As shown in Figure \ref{fig:route_based_fusion}, the process begins with obtaining forecasting results from each TSFM, then extracting encoder embeddings of input time series from any TSFM (with encoder) in the model pool, for example Le-TSFM. Finally, these embeddings are fed into a transformer-based classifier, which assigns appropriate weights to each TSFM's output. Such router network, by assessing the time series hidden characteristics, effectively integrates prediction results from multiple TSFMs together. It can not only intelligently synthesize diverse predictive insights, but also account for the underlying temporal structures of time series. Consequently, this method can enhance the accuracy and robustness of the final forecast. 

\begin{figure}[t]
    \centering
    \includegraphics[width=1\linewidth]{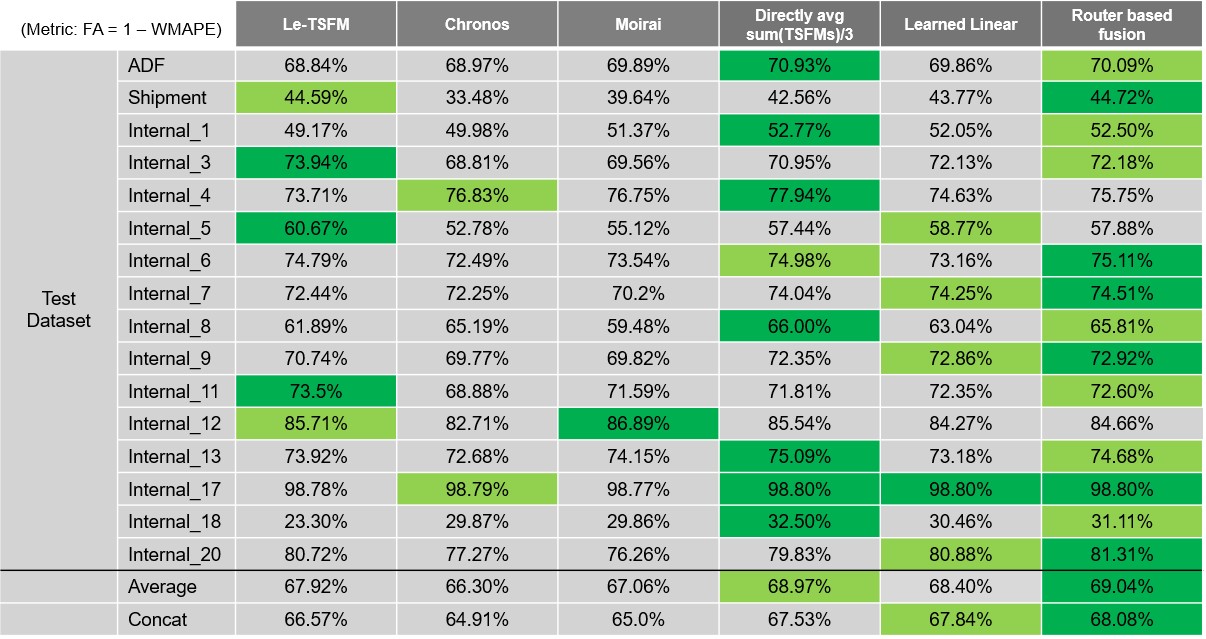}
    \caption{Performance testing results of router-based fusion framework.}
    \label{fig:syn_dataset_vis_fusion}
\end{figure}

We utilized above router network to fuse three TSFMs, including Le-TSFM, Chronos and Moirai. Meanwhile, they are also benchmarks in our experiments, along with a simple averaging fusion approach (Directly avg, i.e. sum(TSFMs)/3) and linear fusion approach (Learned Linear). For the evaluation metric, we applied FA = 1 - WMAPE, where the weights of MAPE were true values of the input time series. Performance testing was conducted on multiple internal Lenovo datasets that were not seen during pre-training of TSFMs. As illustrated in Figure \ref{fig:syn_dataset_vis_fusion}, testing results demonstrated that the forecasting performance achieved through this router network surpasses that of any single TSFM. Furthermore, this approach yields superior and more stable prediction performance compared to either simply averaging fusion approach or a simple trained linear network. Best of all, the accuracy increases 1.5\%.

Even the simplest fusion works, showing the effectiveness of fusion, however, it increases the challenge of subsequent methods. In addition, the effectiveness of fusion mainly relies on the complementary potentials of the different models, while differences of the forecasting capability among TSFMs have not yet been thoroughly explored. With enhanced understanding of the predictive capabilities and boundaries of each TSFM, it can significantly aid in the design of more effective fusion strategies handle more complex application scenario, furthermore, it can also mitigate their weaknesses of individual TSFM.

\subsection{Large and small model coordination}
\begin{figure}[t]
    \centering
    \includegraphics[width=1\linewidth]{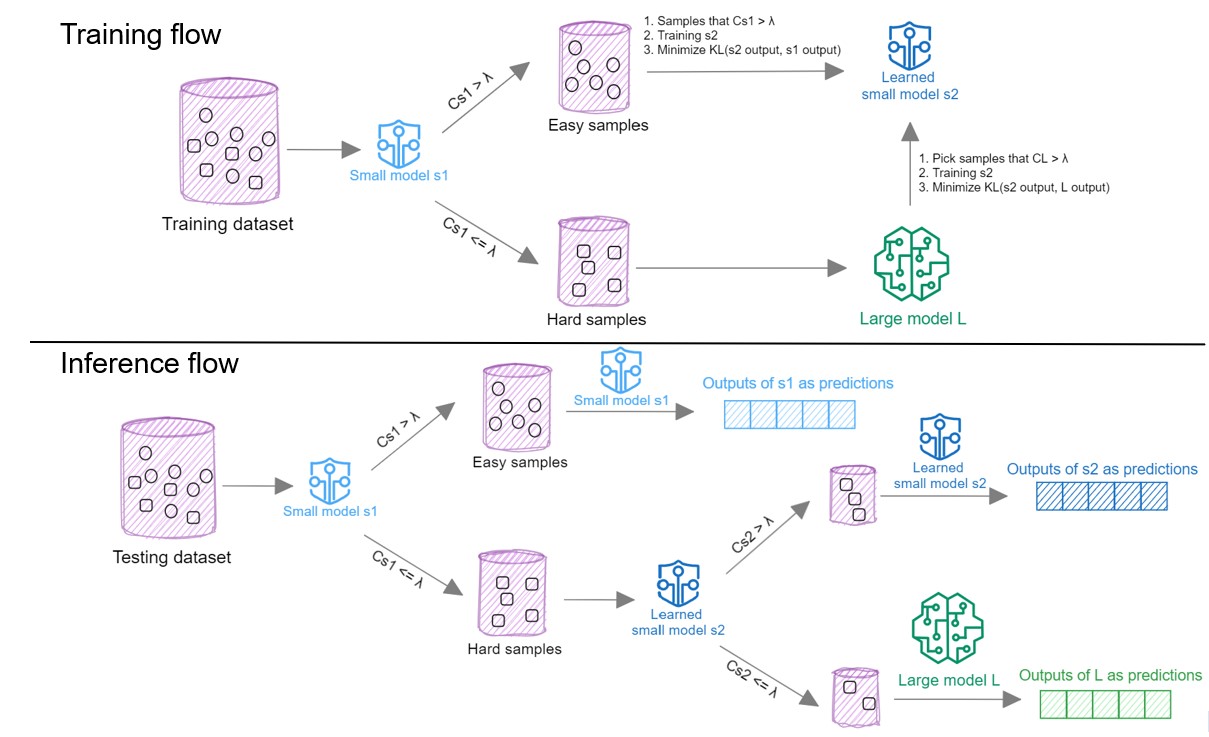}
    \caption{Conceptual Framework of Large and Small Model Coordination}
    \label{fig:model_coordination}
\end{figure}

TSFMs have an excellent generalization ability and can deal with outliers. And SOTA small models, like some deep learning models, can grasp the main patterns of target datasets \cite{xu2023smallmodelsvaluableplugins}. Inspired by confidence-based model selection or fusion strategies \cite{chen2023confidencebasedmodelselectionshortcuts,dong2024datashunt,chen2024improvinglargemodelssmall}, we applied a large and small model coordination framework to take advantage of these complementary strengths.

AS shown in Figure \ref{fig:model_coordination}, the training flow of the proposed coordination framework aimed to distil the forecasting power of Le-TSFM on challenging samples while preserving the performance of a SOTA small model on easy samples. The process unfolded in several steps. Firstly, we trained a small model, denoted as s1, on a target dataset, it could achieve SOTA performance among most small models. Then, each sample is classified based on the forecasting confidence scores. Samples with confidence greater than threshold were labelled as easy samples, while samples with confidence scores less than threshold were designated as hard samples. Next, hard samples are fed into Le-TSFM, samples for which Le-TSFM’s prediction confidence exceeds threshold were labelled as challenging samples. Subsequently, s1 is duplicated to create s2. Finally, the training of s2 aimed to achieve consistency with s1 on easy samples while aligning its forecasting on challenging samples with Le-TSFM.

During inference flow, if s1 confidence for testing samples exceeded the threshold, the forecasting results from s1 were adopted. Samples with s1's confidence below the threshold were passed to model s2. If s2’s confidence surpassed the threshold, their prediction results were adopted. Otherwise, the forecasting results from Le-TSFM were utilized.

This framework can leverage both small and foundation models. However, there are several limitations that warrant further investigations. For examples, the specific characteristics and behaviours of these uncertain samples identified from small models are not fully understood. In addition, it is not guaranteed that the foundation models will always provide more accurate forecasting for samples that are difficult for small models. 

\subsection{Summary}

Model fusion belongs to the promising "learning from model" paradigm, which allows for the
integration of multiple models into a single, cohesive model or framework, leveraging complementary potentials of each time series model to obtain superior performance. Our experiments have demonstrated the effectiveness of the fusion in enhancing predictive performance. In the future, by thoroughly exploring the capabilities and boundaries of time-series models, we will design a more comprehensive fusion framework, achieving an automated and integrated way for model evaluation, model profiling, selection of fusion methods, and model fusion deployment. The "learning from model" paradigm will address complex application scenarios, such as those involving data privacy and difficulties in data collection.

\clearpage
\section{Enterprise Adoption}
\label{sec:adoption}
Aligned closely with the strategic transformation of Lenovo, \leforecast{} has been one of the flywheels that innovates intelligent transformation across multiple business sectors within Lenovo in the last few years. This section offers three examples of business applications pioneered with \leforecast{} that drive efficiency, cost reduction, and sustainability in the operation of global e-Commerce retail, logistics, and Environmental, Social, and Governance (ESG), demonstrating the business value of \leforecast{} for enterprise innovations. 

\subsection{Demand Forecasting}
\begin{figure}[ht]
    \centering
    \includegraphics[width=1\linewidth]{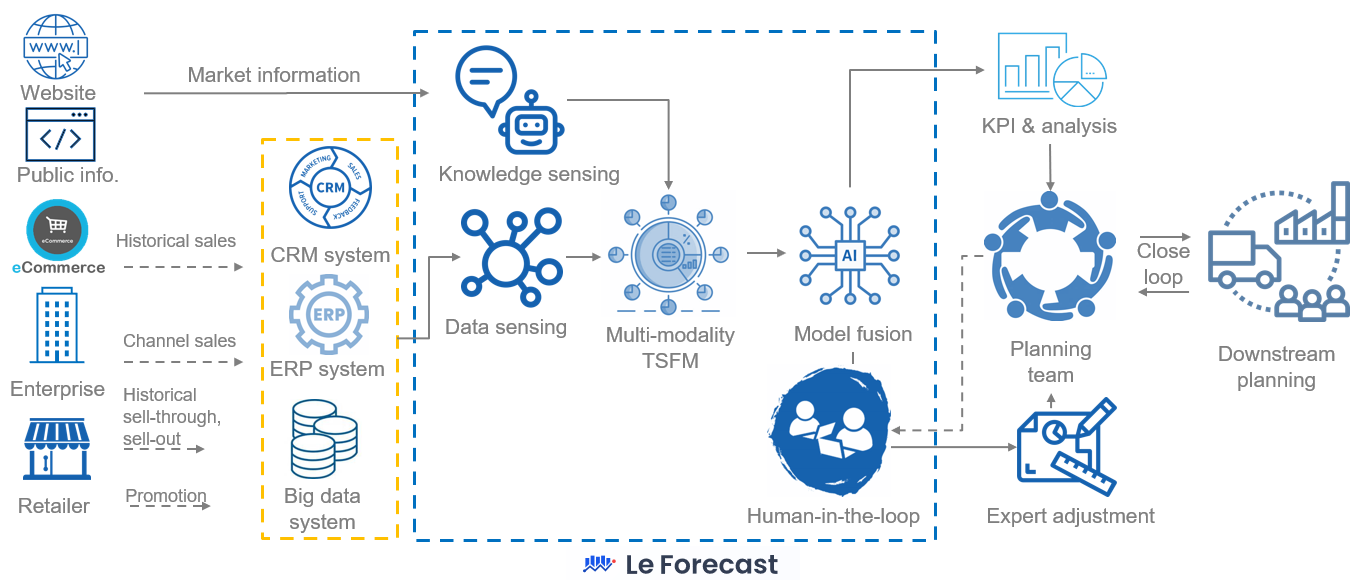}
    \caption{Framework of Demand Forecasting}
    \label{fig:ent_adf}
\end{figure}

Global E-Commerce of Lenovo is the first business domain that adopts \leforecast{} for intelligent transformation. In Lenovo, Global E-Commerce operations, such as inventory management and supply chain planning, are deeply based on accurate forecasts of market demand on digital devices and critical parts for manufacturing. Traditionally, demand forecasts are based on expert knowledge in addition to historical statistical analysis. Thereby, enormous efforts are required to deliver such forecasts of handstands and thousands of forecasting targets, i.e. devices or parts. In addition, forecasting the demands on new products faces a lack of data for reference and solely relies on business knowledge which could differ among experts and locations. \leforecast{} supports the intelligent transformation of these operations with its generalizability and effectiveness in delivering accurate forecasts.

Technically, \leforecast{} first automatically extracts unstructured business knowledge and historical data from multiple sources, respectively, and then forms them as covariate inputs and uses multimodal data fusion before calling zero-shot inference and fusing multiple smaller fine-tuned models to produce hybrid forecasts. Specifically, to align the forecasts with varying business interpretations, \leforecast{} uses the Human-In-The-Loop design to take adjustments from business experts and then improves the adaptation of the forecasts for the planning team. In addition, \leforecast{} also learns from business feedback to optimize its performance systematically through such a loop. Figure \ref{fig:ent_adf} illustrates the detailed framework of our implementation.


Since the adoption of \leforecast{}, Global E-Commerce has achieved significant efficiency gains. It has enhanced supply and manufacturing planning by improving the demand forecast accuracy for customized PC products and their critical components. This has reduced the Order-to-Ship cycle time by 31\%, contributing to a 5\%-7\% year-on-year increase in gross profits. In this case, \leforecast{} has proven its valuable influence on business outcomes, leveraging its technical strengths to empower E-Commerce planning and operations teams.


\subsection{Global Logistics Shipment Forecasting}
Global logistics is another domain leveraging \leforecast{} for intelligent transformation. Lenovo collaborates with multiple third-party freight vendors, requiring advance booking of transportation resources like containers. The underestimation of booking space will introduce additional costs. As a rolling-forecast of shipment volume with accountable fidelity is critical for cost-effective logistics operations, \leforecast{} is proven with its technical capabilities to address these challenges as an adaptive resolution.

Given the volatility of daily shipment data, which includes shipment histories, manufacturing plan, and shipment schedule, \leforecast{} begins by enhancing data quality through outlier detection, pattern recognition, and data imputation. It then evaluates the predictability of various shipment lanes (logistic routes from origins to destinations) by time series analytics. Finally, it leverages model fusion, combining our \letsfm{} framework with classic deep learning models, to generate accurate shipment volume forecasts for each lane.

    


In practice, \leforecast{} has enhanced the efficiency of logistics planning team, with improving forecast accuracy by 20\%. 
Currently, we are scaling this solution to support other industries facing similar logistics challenges, helping them reduce operational costs and improve efficiency.

\subsection{Carbon Emission Forecasting for Lenovo’s Tianjin Factory}

Driven by global ESG goals and carbon neutrality policies, Lenovo aims to reduce carbon emissions from manufacturing. An innovative carbon emission forecasting solution leveraging \leforecast{}, was developed and rolled out to Lenovo's advanced manufacruing campus in Tianjin (Lenovo TJSC), supporting Lenovo’s carbon neutrality factory goals.

One of critical challenges in manufacturing domain is to forecast carbon emissions associated with production, especially Scope 1 \& 2 emissions \cite{EmissionScope2024Mckinsey}. Lenovo has adopted a carbon emission factor approach \cite{EmissionFactors2022CAEP} that predicts emissions by calculating the product of carbon emission factors and energy consumption. Using \leforecast{} integrated \letsfm{} (time series foundation model), the platform predicts carbon emission factors and energy consumption. 
For specific conditions that can comply with \leforecast{}, integrated forecast models are applied properly; for rest of conditions, specific mathematical models are developed to collaborate. 
The fusion of these predictions yields precise forecasts for each type of energy consumption.

This practice represents the first systematic, scientific, and accurate short-term and long-term carbon forecasting for ESG sector in Lenovo China. With over 80\% forecasting accuracy, such solution played a crucial role in Lenovo TJSC’s economy-level carbon neutrality factory certification \footnote{Certification: \url{https://www.cc.cesi.cn/certificate.aspx?type_id=2&keyword=CESI2024ZC10015R1M}} and is widely recognised across the industry.

\subsection{Summary}
The adoption of \leforecast{} across the above three business domains in Lenovo exemplifies the transformative potential of \leforecast{} in driving enterprise intelligence. By delivering accurate demand forecasts, improving the efficiency of logistics operation, and supporting carbon emission reduction initiatives, \leforecast{} not only improved efficiency and reduced costs, but also reinforced Lenovo’s commitment to sustainability and market competitiveness. These use cases underline the scalability and adaptability of \leforecast{}, paving the way for broader implementation across industries. The innovative approach empowered by \leforecast{} demonstrates how intelligent transformation can align technological advancements with practical business outcomes, building a benchmark of operation excellence and AI-powered decision-making.

\clearpage
\section{Strategic Blueprint}
\label{sec:outlook}
\leforecast{} takes a stand on driving enterprise optimisation with advance artificial intelligence in the field of time series, considering the impressive capability of foundation models, the importance of time serires forecasting in diverse business sectors and rapid rising of the Enterprise AI era. To view the potentials and evolving directions of \leforecast{}, this section discusses the landscape of time series foundation models developed over recent years, and highlights critical outlooks of \leforecast{} in advancing fields: multivariate learning, automatic model fusion, and multi-modal forecasting with knowledge sensing.

\subsection{Deviations from Rest TSFMs}
\label{sec:deviations}
Recently, various foundation models have been developed for time series data. With better understanding and choosing the appropriate foundation model, we can more effectively and efficiently leverage their capabilities. In the subsequent sections, we will explore the different foundation models available for time series data analysis.

Time series forecasting is a crucial tool across various domains, from finance to healthcare, enabling informed decision-making based on historical patterns. Advanced foundation models like TimeGPT \cite{garza2023timegpt}, TimesFM \cite{das2023decoder}, Chronos \cite{ansari2024chronos}, Moment \cite{Goswami2024MOMENTAF}, Lag-Llama \cite{rasul2023lagllama}, and Moirai \cite{Woo2024UnifiedTO} offer sophisticated capabilities, leveraging transformer architectures and diverse training datasets for accurate forecasting and analysis. These models provide a glimpse into the future of time series analysis, empowering businesses and researchers with powerful tools to navigate complex data landscapes effectively.

\paragraph{TimeGPT} Developed by Nixtla, TimeGPT is the first foundation model for time series available through an open API. Built on a standard encoder-decoder Transformer architecture with 1B parameters, it is trained on a dataset of over 100 billion points from diverse domains such as finance, healthcare, energy, and IoT sensors. TimeGPT supports zero-shot forecasting across multiple sectors, capturing complex temporal patterns without retraining. Its features include robust uncertainty quantification and adaptability to non-stationary data. With an accessible Python SDK and REST API, TimeGPT democratizes advanced forecasting by significantly reducing complexity while achieving state-of-the-art performance. 

\paragraph{TimesFM} Developed by Google Research, TimesFM is a decoder-only foundation model with 200 million parameters. The model is trained on a dataset of 100 billion real-world time points, encompassing both synthetic and actual data from varied sources such as Google Trends and Wikipedia Pageviews. TimesFM is capable of zero-shot forecasting in multiple sectors, including retail, finance, manufacturing, healthcare, and the natural sciences, across different time granularities. Google intends to release TimesFM on its Google Cloud Vertex AI platform, providing its sophisticated forecasting features to external clients.

\paragraph{Lag-Llama}  Created by researchers from the Université de Montréal, Mila-Québec AI Institute, and McGill University, Lag-Llama is a foundation model designed for univariate probabilistic time series forecasting. Build on the foundation of Llama, the model employs a decoder-only transformer architecture which uses variable sizes time lags and time resolutions for forecasting. The model is trained on diverse time series datasets from several sources across six different groups including energy, transportation, economics, nature, air quality and cloud operations. The model is conveniently accessible through the Huggingface library.

\paragraph{Moirai} Developed by Salesforce AI Research, Moirai is a foundation time series model designed for universal forecasting. Moirai is trained on the Large-scale Open Time Series Archive (LOTSA) dataset, which contains 27 billion observations from nine distinct domains, making it the largest collection of open time series datasets. This diverse dataset allows Moirai to learn from a wide range of time series data, enabling it to handle different forecasting tasks. Moirai uses multiple patch size projection layers to capture temporal patterns across various frequencies. An important aspect of Moirai is to use any-variate attention mechanism, allowing forecasts across any number of variables. The code, model weights, and data associated with Moirai are available in the GitHub repository called “uni2ts“

\paragraph{Chronos} Developed by Amazon, Chronos is a collection of pre-trained probabilistic models for time series forecasting. Built on the T5 transformer architecture, the models use a vocabulary of 4096 tokens and have varying parameters, ranging from 8 million to 710 million. Chronos is pretrained on a vast array of public and synthetic data generated from Gaussian processes. Chronos can be easily integrated into a Python environment and accessed via its API.

\paragraph{Moment} Developed collaboratively by Carnegie Mellon University and the University of Pennsylvania, Moment is a family of open-source foundation time series models. It utilizes variations of T5 architectures, including small, base, and large versions, with the base model incorporating approximately 125 million parameters. The model undergoes pre-training on the extensive “Time-series Pile,” a diverse collection of public time-series data spanning various domains. Unlike many other foundational models, MOMENT is pre-trained on a wide spectrum of tasks, enhancing its effectiveness in applications such as forecasting, classification, anomaly detection, and imputation. The complete Python repository and Jupyter notebook code are publicly accessible for utilizing the model.

Compared to previously introduced foundation models for time series, Le-TSFM incorporates several distinctive features in terms of architectural design, sample input construction, pre-training data, and the pre-training process. First, we employ a Mixture-of-Experts (MoE) architecture, enabling the model to achieve a higher effective parameter count while maintaining the same computational cost during inference. For sample construction, we introduce calendar information as an additional positional encoding input to the model, enhancing its ability to represent time series data. This design is particularly beneficial for zero-shot inference on short sequences and fine-tuning in data scarcity scenarios, allowing the model to capture calendar-related seasonality more effectively. Regarding pre-training data, we address the severe inter-domain imbalance in public datasets by leveraging time series decomposition techniques and general convex padding methods for targeted data augmentation (see Section 2.2). Lastly, to ensure balanced performance across diverse domains, we adopt a domain-robust optimization strategy that dynamically adjusts the weights of samples from different domains during the pre-training process.

\subsection{Outlook on Hybrid Forecasting}
\subsubsection{Multi-variates Learning}
In practical enterprise scenarios, multivariate forecasting is essential, as applications like sales and logistics involve numerous variables (e.g., products, freight routes) requiring simultaneous predictions. While most foundation models in time series forecasting are univariate, practical needs and research trends highlight the necessity of robust multivariate models. Compared to univariate models, multivariate models face higher computational complexity, particularly in Transformer-based models, due to intra- and inter-variable attention mechanisms, emphasizing the need for more efficient attention mechanisms. Additionally, the scarcity of publicly available multivariate datasets and their lack of meaningful variable names or causal knowledge complicate reliable benchmarking and make it unclear whether multivariate models outperform univariate ones.

To tackle these challenges, efficient mechanisms must optimize both intra- and inter-variable attention calculations. Current research suggests strategies during training (e.g., Crossformer \cite{zhang2023crossformer}) or inference (e.g., Partial Attention\cite{lee2024partial}) to alleviate computational burdens. For data limitations, constructing simulated multivariate datasets with predefined correlations and causal relationships can guide models in extracting inter-variable dependencies effectively. These datasets, combined with augmentation of existing datasets, can enable foundation models to achieve stronger zero-shot and fine-tuning performance in multivariate time series forecasting.

\subsubsection{Model Fusion}

Our proposed model fusion framework consists of a model pool, model profiling, fusion of isomorphic models and fusion of heterogeneous models. It allows for the integration of multiple models into a single, cohesive model or framework, leveraging complementary potentials of each time series model to obtain superior performance.

In the future, as the understanding of capabilities/boundaries of model deepens, it is worth to design novel model fusion strategies, inspired by diverse ensemble framework or innovative weight aggregation strategies. Furthermore, a benchmark to evaluate different time series model fusion methods should be considered.



\subsubsection{Multi-modal forecasting with knowledge sensing}

Since Dual-Forecaster is trained with multi-modal datasets including shape-based text descriptions, which makes it unable to directly utilize to provide forecasts given context information. To incorporating agentic sensing result into multi-modal forecasting, we can:
\begin{itemize}
    \item Train Dual-Forecaster from scratch using textual data that contains context information.
    \item Train a light-weight to adapt context information with shape-based text descriptions, thus we can directly use pre-trained Dual-Forecaster for forecasting. 
\end{itemize}

\subsection{Summary}
This section overviews time series foundation models developed over recent years and discusses their unique architectures, training strategies and applications. In line with the rapid evolvement of TSFMs and LMs in academia and industry, this section also introduces the blueprint of \leforecast{} towards reliable, scalable and adaptive hybrid forecasting system. It includes the exploration and development of multivariate learning to addressing challenges in computational complexity and data scarcity for multivariate forecasting, model fusion to combine various models and mitigate risks and errors, as well as multimodal knowledge learning to integrate knowledge sensing ability into forecasting models. Through this blueprint, we foresee the great potentials of \leforecast{} in delivering robust, adaptive, and domain-agnostic solutions, addressing real-world complexities across industries, thereby, empower business in the era of enterprise AI.

\clearpage
\section{Conclusion}
\label{sec:conclusion}
This article discusses the evolving landscape of foundational AI for time series forecasting, exploring recent advances in this domain. 

To address key challenges in enterprise forecasting scenarios, we leveraged advanced techniques and ideas from the era of large-language models. First, we constructed cross-domain real-world and synthetic datasets to pre-train a foundational model with excellent zero-shot forecasting and fine-tuning capabilities, enabling accurate and stable baseline predictions for diverse scenarios. Second, recognizing the difficulty of generalization across scenarios due to varying model assumptions, we implemented a dynamic fusion mechanism between the foundational model and small SOTA models to achieve efficient and robust generalization. Lastly, to incorporate critical internal and external unstructured knowledge for enhanced forecasting, we integrated agent-based market sensing techniques and multimodal forecasting methods, enabling effective acquisition, extraction and utilization of diverse forms of knowledge. Combining all those features, the Le-Forecast platform demonstrates strengths in forecast accuracy and scenario generalization. This platform holds substantial technological and commercial value within enterprise-level intelligent applications, providing a solid foundation for developing enterprise-wide planning and operational optimization solutions in the future.

\clearpage
\renewcommand{\thesection}{}
\section*{Acknowledgements}
\addcontentsline{toc}{section}{Acknowledgements}

This work would not have been possible without the invaluable contributions and support of the entire Smart Data team at the Lenovo Research AI Lab. We also extend our sincere appreciation to the individual members whose expertise, dedication, and innovative thinking have significantly shaped this research.

Zheng Tan designed the initial transformer-based framework for the time series foundation model and led efforts to refine its pre-training strategy while introducing multivariate forecasting with Xinyuan Tian and anomaly detection. Guanyu Zhang contributed to the data diffusion methodologies, improving the integration of diverse time-series data sources for a more robust foundation model. Wenfa Wu expanded the framework with multimodal forecasting, integrating textual insights to enhance pattern recognition and humanized interactive forecasting. 

Yiwen Nie developed a unique knowledge-sensing capability, enabling hybrid forecasting models with Haipeng Jiang to incorporate signals beyond traditional time-series data. Yingzheng Ma pioneered the real-world application of models in product demand forecasting, logistics shipment prediction, and long-term carbon emission forecasting, generating critical feedback for optimization. Kailin Gao played a key role in model fusion techniques, integrating statistical models with machine learning approaches to achieve seamless collaboration among different scaled models. 

Finally, Mengya Liu facilitated collaboration and conducted structured surveys, refining the architecture, and providing insightful summaries with Hongsheng Qi that strengthened both theoretical foundations and practical applications.
In addition, we acknowledge Jiang Tian, Peng Wang, Yao Meng, Qigang Wang, Ming He, Zhongchao Shi, Yangzhou Du, the colleagues of Lenovo Research AI Lab for fruitful discussions, offering valuable suggestions and expertise during our research, development and experiments on the models and system. Especially we appreciate much to Jianping Fan, the leader of the Lab for his enlightening guidance as a cornerstone to explore the innovative technical methodologies featured in this paper. Specifically, this research introduces significant advancements in time-series forecasting through the development of a hybrid forecast system that integrates multiple analytical paradigms. By combining different scale machine learning models with transformer-based foundation model techniques, we have established a novel framework that enhances accuracy and adaptability in predictive analysis.

The paper further emphasizes the development of a robust system architecture designed to address the challenges of diverse and uncertain estimations. This architecture not only streamlines the integration of heterogeneous data sources but also ensures scalability and resilience in dynamic environments. Using cutting-edge technologies and design principles, this system bridges theoretical exploration with practical applications, providing a comprehensive tool to tackle complex forecasting problems in real industry use cases.

Furthermore, we acknowledge the critical role that a diverse team plays in ensuring the relevance and applicability of the proposed methodologies. Their efforts in architecture designing, software engineering, and use cases testing relevant new models to build the system have reinforced its potential to empower decision-making processes across various domains. Their dedication has been pivotal in the transformation of innovative ideas into actionable solutions.


\clearpage
\bibliography{literature/references}
\bibliographystyle{apacite}


\appendix
\chapter*{Appendix}
\renewcommand{\thesection}{\Alph{section}}

\clearpage
\section{Additional Results}
\label{sec:appendix-c}


\includepdf[scale=0.75, pages=1, pagecommand=\subsection{Agentic Sensing Result Sample}]{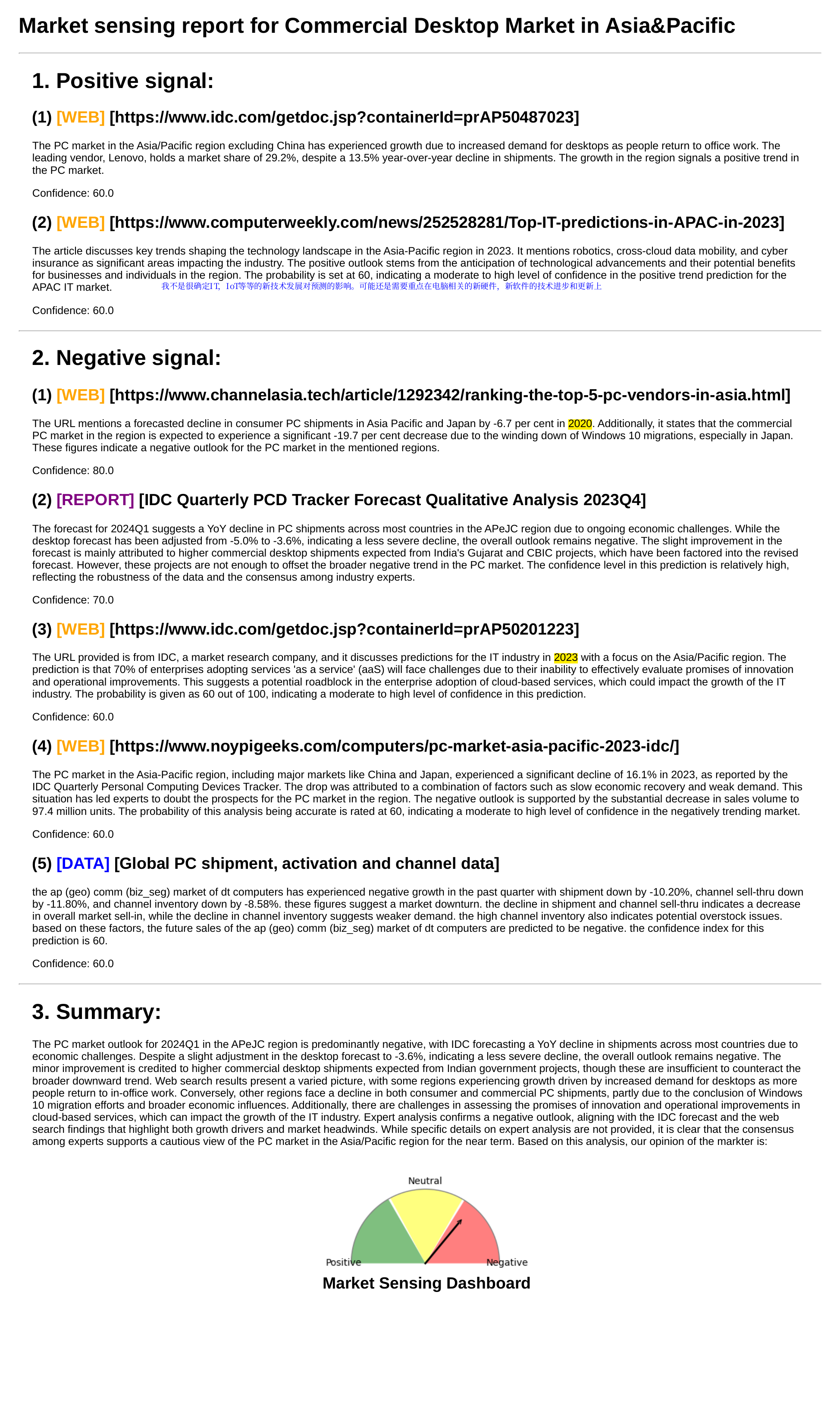}
\label{sec:appendix-c-agenticsensing}


\clearpage
\section{Prompts}
\label{sec:appendix-b}

\subsection{Web Search for News Sensor}
\label{sec:appendix-b-websearch}

\begin{mainbox}{Web Search Prompt} 
    Your role is as an information analyst for the PC market: \\
    Here are some tips to help you make your decision:
    \begin{enumerate}
        \item The release of powerful new processors could boost PC sales. There have been reports that serious security vulnerabilities or performance issues could lead to a drop in PC sales.
        \item The continuing trend of remote working and online education has increased the demand for personal computers in homes and educational institutions, driving sales. The popularity and improved performance of mobile devices such as smartphones and tablets has reduced the demand for PCS, which could lead to a decline in sales.
        \item A stable supply chain and improved production efficiency enable PC manufacturers to meet market demand, potentially boosting sales growth. Supply chain disruptions, such as natural disasters or factory shutdowns, lead to product shortages, which can limit availability and lead to lower sales.
        \item Falling revenues and declining demand for personal computers could tilt the decision negative
    \end{enumerate}
    
    <query>: use search tool to Search for news related to {businesssegment} {productsegment}, the region is {region}\\       
        the result should reply as follow structured type: \\
        \small \textit{\{ \\ 
            \-\hspace{0.5cm} "each" :$[$ \\ 
                \-\hspace{1cm} $[$"url":"url", "result": "positive/negative", "probability": "<probability>", "reason":"<reason>" $]$, \\
                \-\hspace{1cm} $[$"url":"url", "result": "positive/negative", "probability": "<probability>", "reason":"<reason>" $]$, \\
                \-\hspace{1cm} $[$"url":"url", "result": "positive/negative", "probability": "<probability>", "reason":"<reason>" $]$, \\
            \-\hspace{0.5cm}$]$, \\
            \-\hspace{0.5cm} "overall" : $[$ "overall result", "positive/negative","overall probability": "average probability of all the <probability>", "overall reason": "overall reason of all <reason>" $]$\\
        \}
        }
\normalsize
        \begin{enumerate}
            \item "each" is the analysis of each searched urls, \\ 
             "overall": an overall result, overall probability, and reason of the searched informations.

            \item result: Give positive/negative market sentiment for each url, do not give other type

            \item probability: Confidence level of the "result," ranging from 1 to 100
                \begin{itemize}
                    \item 1 to 49 indicates low confidence
                    \item 50 to 100 indicates high confidence
                \end{itemize}
            
            \item reason:
                 \begin{itemize}
                    \item Summarize the url information
                    \item Provide a detailed summary and reason of the "result" and "probability"
                \end{itemize}            
        \end{enumerate}

\end{mainbox}

\subsection{Data Analysis for Signal Sensor}
\label{sec:appendix-b-data}

\begin{mainbox}{Data Analysis Prompt} 
    As a market analyst, your task is to analyse the performance data of \colorbox{cyan}{AP} (GEO), \colorbox{cyan}{AP} (Bizseg) market of \colorbox{cyan}{AP} computers and provide a future sales prediction. You are provided with detailed data points and logic to help with the analysis. The data can be found below, surrounded by input tags. \\
    
    <input>: For the \colorbox{cyan}{AP} (GEO), \colorbox{cyan}{AP} (Bizseg) market of \colorbox{cyan}{AP} computers, the past quarter year-over-year changes are as follows: 
        Shipment is {row['Shipment']}, \\
        Channel Sell Thru is {row['Channel Sell Thru']}, \\
        Channel Inventory is {row['Channel Inventory']}, \\
        Also, here are some basic logic to analysis the market:
        \begin{enumerate}
            \item Shipment: 
            \begin{enumerate}
                \item Growth in shipments indicates an increase in overall market sell-in. A decline in shipments indicates a market downturn. 
                \item Shipment Data: This refers to the sell-in data for Lenovo, representing the number of units shipped from the company to distributors.
            \end{enumerate}

            \item Channel Sell-Through: 
            \begin{enumerate}
                \item Growth in channel sell-through indicates an increase in overall market sell-in. A decline indicates a market downturn.
                \item Channel Sell-Through Data: This measures the movement of PCs from distributors to retailers across the entire market.
            \end{enumerate}

            \item Channel Inventory: 
            \begin{enumerate}
                \item High Channel Inventory: Indicates that PCs are not being sold to end customers efficiently, potentially leading to overstock.
                \item Low Channel Inventory: Indicates a potential shortage in PC supply, suggesting strong demand.
            \end{enumerate}

        \item* Summary:
            \begin{enumerate}
                \item Growth in shipments, or channel sell-through suggests an overall increase in market sell-in, while declines indicate market downturns.
                \item Data from shipment is Lenovo company-specific, while channel data reflect market-wide trends.
            \end{enumerate}
        \end{enumerate}
        </input> \\
        
        Based on these data, provide an analysis of the future sales of \colorbox{cyan}{AP} (GEO), \colorbox{cyan}{AP} (Bizseg) market of \colorbox{cyan}{AP} computers. Predict whether the future sales will be positive, negative, or neutral. Provide a confidence index for your prediction ranging from 1 to 100, where 1 to 49 indicates low confidence and 50 to 100 indicates high confidence.
        
        Return output as a string in the following JSON format and do not include: \\
        \fcolorbox{white}{black!30!white}{analysis:<analysis>,prediction: <prediction>, confidence:<confidence>}   \\

        Let' s work this out in a step-by-step way to be sure we have the right answer.
\end{mainbox}

\subsection{Knowledge integration for Analyst module}
\label{sec:appendix-b-report}
\begin{mainbox}{Knowledge integration prompt} 
    As an Analyst, you got the research report analysis with + final['Report Analysis']+You also got the web search analysis"  + final['News Analysis'] + You also got the  data analysis"  + final['Data Analysis'].\\
    Now you can make a summary based on the analysis from three knowledge sources about PC market, The first paragraph is report summary, and the second paragraph is news summary, the third part is the data summary. \\
    Finally, you can make a comprehensive summary based on the above summaries to output the trend prediction.

\end{mainbox}




\end{document}